\documentclass[10pt, a4paper]{article}
\usepackage{microtype}
\usepackage{graphicx}
\usepackage{lmodern}
\usepackage{booktabs} 
\usepackage{hyperref}
\usepackage{amsfonts}
\usepackage{mathtools}
\usepackage{amssymb}
\usepackage[table]{xcolor}
\usepackage{dsfont}
\usepackage[utf8]{inputenc} 
\usepackage[T1]{fontenc}    
\usepackage{url}            
\usepackage{nicefrac}       
\usepackage{microtype}      
\usepackage{comment}
\usepackage{multirow}
\usepackage{algorithm}
\usepackage{algorithmicx}
\usepackage{etoolbox}
\usepackage{algpseudocode}
\usepackage{wrapfig}
\PassOptionsToPackage{square,numbers}{natbib}
\usepackage{natbib}
\usepackage{caption}
\usepackage{mwe}
\usepackage{hyphenat}
\usepackage{subfig}

\allowdisplaybreaks

\usepackage{setspace}
\let\Algorithm\algorithm
\renewcommand\algorithm[1][]{\Algorithm[#1]\setstretch{1}}

\definecolor{Gray}{gray}{0.9}

\newcommand{\transpose}{^\mathsf{T}}

\newcommand{\argmax}{\mathop{\rm argmax}}
\newcommand{\trace}{\mathop{\rm Tr}}

\newcommand{\Probability}{\mathbb{P}}

\def\blambda{\boldsymbol{\lambda}}
\def\bpi{\boldsymbol{\pi}}
\def\bgamma{\boldsymbol{\gamma}}
\def\bSigma{\mathbf{\Sigma}}
\def\bmean{\mathbf{m}}

\def\bA{\mathbf{A}}
\def\bC{\mathbf{C}}

\def\bT{\mathbf{T}}

\def\bX{\mathbf{X}}

\def\bsx{\mathbf{x}}

\def\cN{\mathcal{N}}

\def\Si{\mathbb{S}}
\def\reals{\mathbb{R}}

\title{Local Bures-Wasserstein Transport: A Practical and Fast Mapping Approximation}

\begin{document}

\author{%
	Andrés Hoyos-Idrobo\\
	Rakuten Institute of Technology, Paris}

\maketitle

\vspace{-2mm}
\begin{abstract}
%
Optimal transport (OT)-based methods have a wide range of applications and have attracted 
a tremendous amount of attention in recent years.
However, most of the computational approaches of OT
do not learn the underlying transport map.
Although some algorithms have been proposed to learn this map, they rely on kernel-based methods,
 which makes them prohibitively slow when the number of samples increases.
Here, we propose a way to learn an approximate transport map and a 
parametric approximation of the Wasserstein barycenter.
%
%
%
We build an approximated transport mapping by leveraging the closed-form 
of Gaussian (Bures-Wasserstein) transport; 
we compute local transport plans between matched pairs of the Gaussian
 components of each density. 
The learned map generalizes to out-of-sample examples.
We provide experimental results on simulated and real data, 
comparing our proposed method with other mapping estimation algorithms.
Preliminary experiments suggest that our proposed method is not only faster, 
with a factor 80 overall running time,
but it also requires fewer components than state-of-the-art methods 
to recover the support of the barycenter.
From a practical standpoint, it is straightforward to implement and 
can be used with a conventional machine learning pipeline.
%
\end{abstract}

\section{Introduction \label{introduction}}
\vspace{-1mm}
In recent years, Optimal Transport-based algorithms (OT)\cite{peyre2019computational} 
are increasingly attracting the machine learning community.
OT  leverages useful information about the nature of 
a problem by encoding the geometry of the underlying space through a cost metric.
As OT measures distances between probability distributions, it  
is generally used as an optimization loss 
\cite{arjovsky2017wasserstein, flamary2018wasserstein, genevay2018learning, tolstikhin2017wasserstein}.
OT aims at finding a probabilistic matrix that couples two distributions;
 this matrix is also called a transportation matrix. 
However, we can only use this matrix with the data used to learn it. 
Thus, one has to recompute an OT problem to handle new samples. 
Hence, it is prohibitive for some applications 
where the OT problem is not used as a distance to be minimized,
for instance, transport out-of-sample data into a fixed distribution \cite{courty2017optimal}.
Another critical application is computing a weighted mean in a Wasserstein space.
This weighted mean is known as the Wasserstein barycenter, 
which has numerous applications in unsupervised learning.
It is used to represent 
the input data \cite{schmitz2018wasserstein}, to cluster densities \cite{ye2017fast},  
to analyze shapes \cite{rabin2011wasserstein}, to build fair classifiers \cite{del2018obtaining}, 
among other applications.

To handle unseen data, 
\cite{perrot2016mapping} proposes to learn a transport map.
%
However, this approach relies on kernel-based methods, which makes it slow when 
the number of samples increases.
%
%
To approximate the barycenter,  
\cite{bonneel2011displacement} proposes to decompose the input data into sums of 
radial basis functions and perform partial transport between matched pairs of functions
 coming from each distribution.
However, it is not clear what is the influence of noise in this approximation as it handles 
the positive and negative parts of the data independently.
Additionally, the extension of this approach to out-of-sample data is not explored.

\paragraph{Contributions.}
%
We propose to use a local approximation of the Gaussian (Bures-Wasserstein) transport 
to speed up learning mappings and barycenters that can be applied to unseen data. 
This approach assumes that the densities involved in the transport problem are highly concentrated 
in a few clusters/components.
%
%
This idea is reminiscent of the computation of barycenters by partial transport \cite{bonneel2011displacement}. 
%
The key point is to express each density as a mixture of Gaussians; contrary to \cite{bonneel2011displacement}, 
it is no longer necessary to split the input data into its positive and negative parts.
Then, we can use the means and covariances of these Gaussian functions to approximate 
the transport map as well as the barycenter. 
To do so, we use these means as a new spatial reference. 
Then, we match pairs of means of different distributions and solve OT problems for each matched pair.
%
%

The proposed method unifies the learning of the barycenter and transport map.
We provide some empirical evidence for the usefulness of our approach in two tasks: 
\emph{i)} interpolating of clouds of points, and 
\emph{ii)}  building fair classifiers.
%
%
Through experiments on public datasets, we show 
the improvements brought by our scheme with respect to standard mapping estimation methods.
Our approach can be easily implemented and combined with conventional machine learning pipelines.


\paragraph{Notation.} Column vectors are denoted as bold lower-case, e.g, $\mathbf{a}$.
Matrices are written using bold capital letters, e.g., $\mathbf{A}$.
Let $\|\cdot\|_F$ be the Frobenius norm of a matrix, 
$\trace(\cdot)$ is the trace of a matrix, and
%
$\mathbf{A}\transpose$ is the transpose of $\mathbf{A}$.
We denote the $(m - 1)$-dimensional probability simplex by 
$\Delta^m \coloneqq \{\mathbf{y} \in \mathbb{R}_{+}^m: \|\mathbf{y}\|_1 = 1\}$.
Let $n$ be the number of data points and $[n]$ denotes $ \{1, \ldots, n\}$. 
\section{Background: Gaussian optimal transport}
%
We will restrict ourselves to the discrete setting 
(i.e., 
histograms);
thus, $\mu \in \Delta^{p}$ and $\nu \in \Delta^{q}$.
The cost matrix $\mathbf{C} \in \mathbb{R}^{p\times q}$ represents
the cost function (e.g., the squared Euclidean distance),
which contains the cost of transportation between any two locations in the discretized grid.
The OT distance between two histograms $W(\mu, \nu) $ is the solution of 
the discretized Monge-Kantorovich problem \cite{villani2008optimal}:
\begin{equation}
\label{eq:kantorovich}
	W^2(\mu, \nu) = \min_{\mathbf{T} \in \Pi(\mu, \nu)} \trace(\mathbf{T}\transpose\, \mathbf{C}),
\end{equation}
where $\mathbf{T}: \reals^p \rightarrow \reals^q$ is the transport plan, $\Pi(\mu, \nu)$ 
denotes the set of admissible couplings between $\mu$ and $\nu$, that is, 
the set of matrices with rows summing to $\mu$ and columns to $\nu$.

However, computing Eq.(\ref{eq:kantorovich}) for other distributions than Gaussians 
requires to solve a linear programming problem  that grows at least $O(p^3 \log p)$ \cite{peyre2019computational} for histograms of dimension $p$.
%
We can constrain the entropy of the transport plan $\mathbf{T}$ to alleviate this issue \cite{cuturi2013sinkhorn}. 

\paragraph{Bures-Wasserstein transport.}
The OT distance in Eq.(\ref{eq:kantorovich}) can be calculated in closed-form  
when $\mu$ and $\nu$ are Gaussians measures, 
 as demonstrated by \cite{givens1984class}  and discussed by \cite{takatsu2011wasserstein}.
This corresponds to the Bures-Wasserstein distance \cite{bhatia2018bures} between 
$\mu  \sim \mathcal{N}(\mathbf{m}_1,  \bSigma_1)$ and $\nu \sim \mathcal{N}(\mathbf{m}_2,  \bSigma_2)$:
\begin{equation}
		W_2^2(\mu, \nu)  = \|\bmean_1 - \bmean_2\|_2^2 + \trace(\bSigma_1 + \bSigma_2) 	- 2\, 
		\trace\left[\left(\bSigma_1^{1/2}\, \bSigma_2\, \bSigma_1^{1/2}\right)^{1/2}\right],
\end{equation}
where 
$\bmean_i \in \reals^{p}$ and  $\bSigma_i \in \reals_{+}^{p \times p}$ for $i \in [2]$ are the mean vectors and the covariance matrices, respectively.
$W_2^2(\cdot, \cdot)$ uses the squared Euclidean distance as transport cost.
For zero-mean Gaussian random variables, the Wasserstein distance reduces to Frobenius of the covariance roots \cite{masarotto2018procrustes}, and the transport map is: 
%
%
\begin{equation}
\label{eq:transport_plan_covariances}
\bT_{\bSigma_1}^{\bSigma_2} =  \bSigma_1^{-1/2}\, \left( \bSigma_1^{1/2} \bSigma_2\, \bSigma_1^{1/2}\right)^{1/2} \bSigma_1^{-1/2},
\end{equation}
where $\bT_{\bSigma_1}^{\bSigma_2}$ is the transport map from $\bSigma_1$ to $\bSigma_2$.
In general, the Gaussian transport $\bT: \reals^p \rightarrow \reals^p$ is 
\begin{equation}
\label{eq:transport_plan}
\bT(\mathbf{x}) =	\bmean_2 + \bT_{\bSigma_1}^{\bSigma_2}  (\mathbf{x} - \bmean_1).
\end{equation}

\paragraph{Bures-Wasserstein  barycenter.}
The Wasserstein barycenter problem attempts to summarize a collection 
of probability distributions by taking their weighted average with respect to 
the Wasserstein metric.
In particular, the Wasserstein barycenter of $l$  Normal distributions 
$\cN(\bmean_0, \bSigma_0), \ldots,  \cN(\bmean_l, \bSigma_l)$,  
with weights $\blambda \in \Delta^l$ 
is the solution of the 
minimization problem
\begin{equation}
\label{eq:gaussian_barycenter}
\min\limits_{\bar{\bmean} \in \reals^{p}, \bar{\bSigma} \in \Si_+} \sum\limits_{i=1}^{l} \blambda_i\, 
W_2^2(\cN(\bar{\bmean}, \bar{\bSigma}), \cN(\bmean_i, \bSigma_i)).
\end{equation}

The barycenter of Gaussians is still Gaussian, and its mean $\bar{\bmean}$ can be computed directly as 
$\bar{\bmean} = \sum_{i=0}^{l} \blambda_i\, \mathbf{m}_i$.
However, its covariance $\bar{\bSigma} $ is the solution of the following matrix equation
\begin{equation}
\label{eq:barycenter_fixed_point}
\bar{\bSigma} = \sum_{i=1}^{l} \blambda_i \left(\bar{\bSigma}^{1/2}\, \bSigma_i\, 
\bar{\bSigma}^{1/2}\right)^{1/2},
\end{equation}
which has not a closed-form solution for $l > 2$.

\begin{wrapfigure}{R}{0.52\textwidth}
	\vspace{-25pt}
	\begin{minipage}[t]{0.48\textwidth}
		\begin{algorithm}[H]\small
			\caption{Barycenter of covariances \cite{masarotto2018procrustes}}
			\label{alg:gaussian_barycenter}
			\begin{algorithmic}[1]
				\Require{$\bSigma_i \in \reals^{p\times p}$, $i \in [l]$, and $\blambda \in \Delta^{l}$.}
				\While{Stopping criterion is not met} 
				\State Compute $\mathbf{T}_{\bSigma_i}^{\bar{\bSigma}^t}$ for each $i \in [l]$ using Eq.(\ref{eq:transport_plan_covariances})
				\State Linear averaging,  $\mathbf{T} = \sum\limits_{i=1}^{l} \blambda_i\, \mathbf{T}_{\bSigma_i}^{\hat{\bSigma}^t}$
				\State Retraction, $\bar{\bSigma}^{t+1} = \mathbf{T} \bar{\bSigma}^t \mathbf{T} $
				\State Set $t = t + 1$
				\EndWhile
				\State\Return{Barycenter $\bar{\bSigma}^{t+1}$}
			\end{algorithmic}
		\end{algorithm}
	\end{minipage}	
	\vspace{-30pt}
\end{wrapfigure}
The algorithm \ref{alg:gaussian_barycenter}\cite{masarotto2018procrustes} 
finds a fixed point of Eq.(\ref{eq:barycenter_fixed_point}). 
It initializes the barycenter with a positive definite matrix. 
Then, it lifts all observation to the tangent space at the initial guess via log map
 and performs a linear averaging.
This average is then retracted onto the manifold via the exponential map, 
providing a new guess. 
The algorithm iterates until it reaches the stop condition (e.g., number of iterations).
\section{The model}
%
We are interested in a fast approximation of a transport map 
that generalizes to out-of-samples data.
We assume the data lives in a low-dimensional feature space, 
and a few Gaussians concentrate the samples.
Thus, we use mixtures of Gaussians to approximate independently the 
densities $\mu$ and $\nu$  involved in Eq.(\ref{eq:kantorovich}).  
%
%
%
Then, we compute the pairwise transport plans (see Eq.\ref{eq:transport_plan}) 
at the component level.
%
%
We refer to this approach as local Bures-Wasserstein (L-BW) transport.

\paragraph{Gaussian Mixture Model (GMM).}
We model the density $\mu$ as a weighted average of $k$ Gaussian functions, 
where each Gaussian or component represents a subpopulation of the data. 
The parameters of each component $i \in [k]$ are the mean vector $\bmean_i^\mu \in \reals^{p}$ and 
covariance matrix $\bSigma_i^\mu \in \reals_{+}^{p \times p}$.
%
%
\begin{equation}
\label{eq:gmm}
\Probability_k
[\textbf{x}] =\sum_{j=1}^{k} \bpi_j^\mu\, \phi(\bsx;\, \bmean_j^\mu, \bSigma_j^\mu),
\end{equation}
where 
$\bpi^\mu \in \Delta^k$ is the mixing proportions,
and $\phi(\cdot;\, \bmean_j, \bSigma_j)$ denotes a $p$-dimensional 
Normal density with mean vector $\bmean_j$ and covariance matrix $\bSigma_j$.
%
We usually use the Expectation-Maximization (EM) algorithm to fit a GMM \cite{friedman2001elements}.
This algorithm alternates between estimating the probability to assign a sample $\bsx$ to component $j$, 
$\bgamma_j(\bsx) \in \Delta^k$, and calculating the parameters of model. 
We denote $\delta(\bsx)$ the (hard) assignment of sample $\bsx$ to a component,
%
%
\begin{equation}
	\delta(\bsx) = \argmax_{j \in [k]} \bgamma_j(\bsx).
\end{equation}

\subsection{Leveraging Bures-Wasserstein transport. \label{sub:leveraging_BW}}
%
For simplicity, we describe our approach for two densities/groups.
We depict how to extend it to more groups at the end of this section.
Let $g \in \{a, b\}$ be the group indicator, and 
$\bX^g \in \reals^{|q| \times p}$ be the set of samples $\bsx$ that 
belong to the group $g$, where $|g|$ denotes the number of samples of group $g$.

\paragraph{Getting the parameters.}
We aim to leverage the closed-form of the Bures-Wasserstein transport (see Eq.(\ref{eq:transport_plan}))
to learn an approximated transport map. 
To do so, for each $g \in \{a, b\}$,  we use a $k$ mixture of Gaussians to approximate the density of $\bX^g$.
We use the means  $\{\bmean_i^g\}_{i=1}^k$ as a rough approximation of the geometry. 
Then, we solve $k$ pairwise Bures-Wasserstein OT problems.
We select pairs because the OT of Gaussians is still Gaussian, 
removing important geometrical information.

To select which components from each group to pair, 
we solve a linear allocation problem 
\cite{burkard2009assignment} 
of their mean vectors.
In particular, we use the Hungarian algorithm \cite{munkres1957algorithms} 
to match these values.
This algorithm has a time complexity $O(k^3)$,  where $k \ll \min(|a|, |b|)$, and 
solves an OT problem for uniform distributions of the same size \cite{peyre2019computational}.
We set the transport cost to $\bC_{i, j} = \|\bmean_i^a - \bmean_j^b\|^2, \, \bC \in \reals^{k \times k}$. 
We denote $\bA \in [0, 1]^{k \times k}$ a matrix that encodes the pairwise matching of components.
Thus, for $(i, j) \in [k]^2$, $\bA_{i, j} = 1$ if $\bmean_i^a$  match $\bmean_j^b$ and $0$ otherwise.

From now on, we rely on $\bA$, $\left\{(\bmean_l^g, \bSigma_l^g)\right\}_{l=1}^k$ and $\delta(\cdot)^g$ to 
approximate the barycenter and the transport map.
The algorithm \ref{alg:gmm_wasserstein} presents a pseudo-code of learning these parameters.
%
\begin{algorithm}[H]\small
	\caption{Learning parameters of L-BW transport.}
	\label{alg:gmm_wasserstein}
	\begin{algorithmic}[1]
		\Require{$\bX^g \in \reals^{|g| \times p}$ for $g \in \{a, b\}$, and a number $k$ of components.}
		
		\State  Estimate parameters of GMM from samples $\bX^g$: 
		Get 
		$\left\{(\bmean_l^g, \bSigma_l^g)\right\}_{l=1}^k$ and $\delta^g(\cdot)$ for $g \in [a, b]$.
		
		\State  Get matching matrix $\bA$: 
		Set $\bC_{i, j} =  \| \bmean_i^a - \bmean_j^b\|^2$ for $(i, j) \in [k]^2$ and
		use the Hungarian algorithm \cite{munkres1957algorithms}.
		
		\State\Return{matching matrix $\bA$,  and GMM parameters $\left\{(\bmean_l^g, \bSigma_l^g)\right\}_{l=1}^k$} and $\delta^g(\cdot)$.
	\end{algorithmic}
\end{algorithm}

We can extend this approach for a number 
$s$ of groups by sequentially computing 
one reference group and the other populations (a.k.a. one-vs-rest approach).
This has a time complexity of $O(s \,k^3)$. 
%

\paragraph{Approximate a transport map.}
To transport sample $\bsx$ from a distribution $a$ to another $b$,  
we use the algorithm \ref{alg:gmm_wasserstein} and apply Eq.(\ref{eq:transport_plan})
for each matched pair of components, as follows 
\begin{equation}
\label{eq:local_affine_transport}
\begin{split}
\hat{\bT}_{a}^{b}(\bsx) & = \sum_{(i, j) \in [k]^2} \bA_{i, j} \left[\bmean_j^b +
\Gamma(\bsx, i) \right],\\
\text{where }\Gamma(\bsx, i) & = \begin{cases}
\bT_{\bSigma_i^a}^{\bSigma_j^b}  (\bsx - \bmean_i^a) & \delta^a(\bsx) = i,\\
0 & \text{otherwise},
\end{cases}\\
\end{split}
\end{equation}
%
where $\hat{\bT}_{a}^{b}(\bsx)$ denotes the approximated transport map of $\bsx$ from 
the population $a$ to population $b$.
In a nutshell, our algorithm performs pairwise transport between components of two distributions, $a$ and $b$.
First, it uses $\delta^a(\bsx)$ to assign a sample $\bsx$ to a $i$ subpopulation of $a$, which has parameters $\left(\bmean_i^a, \bSigma_i^a\right)$.
Then, it matches  the $i$-Gaussian of group $a$ with a $j$-component of group $b$,
parametrized by  $\left(\bmean_j^b, \bSigma_j^b\right)$.
Next, it computes the Bures-Wasserstein transport of these two Gaussians.
Finally, It repeats for all pairs of matched components\footnote{
	We can use our approach with Spark \cite{zaharia2016apache} for large-scale data applications; the GMM is part of SparkML.
}.

\begin{figure}
\centering
\small
\subfloat[
Input data.
]{\includegraphics[width=0.2\linewidth, trim={0mm 0mm 0mm 5mm}, clip]{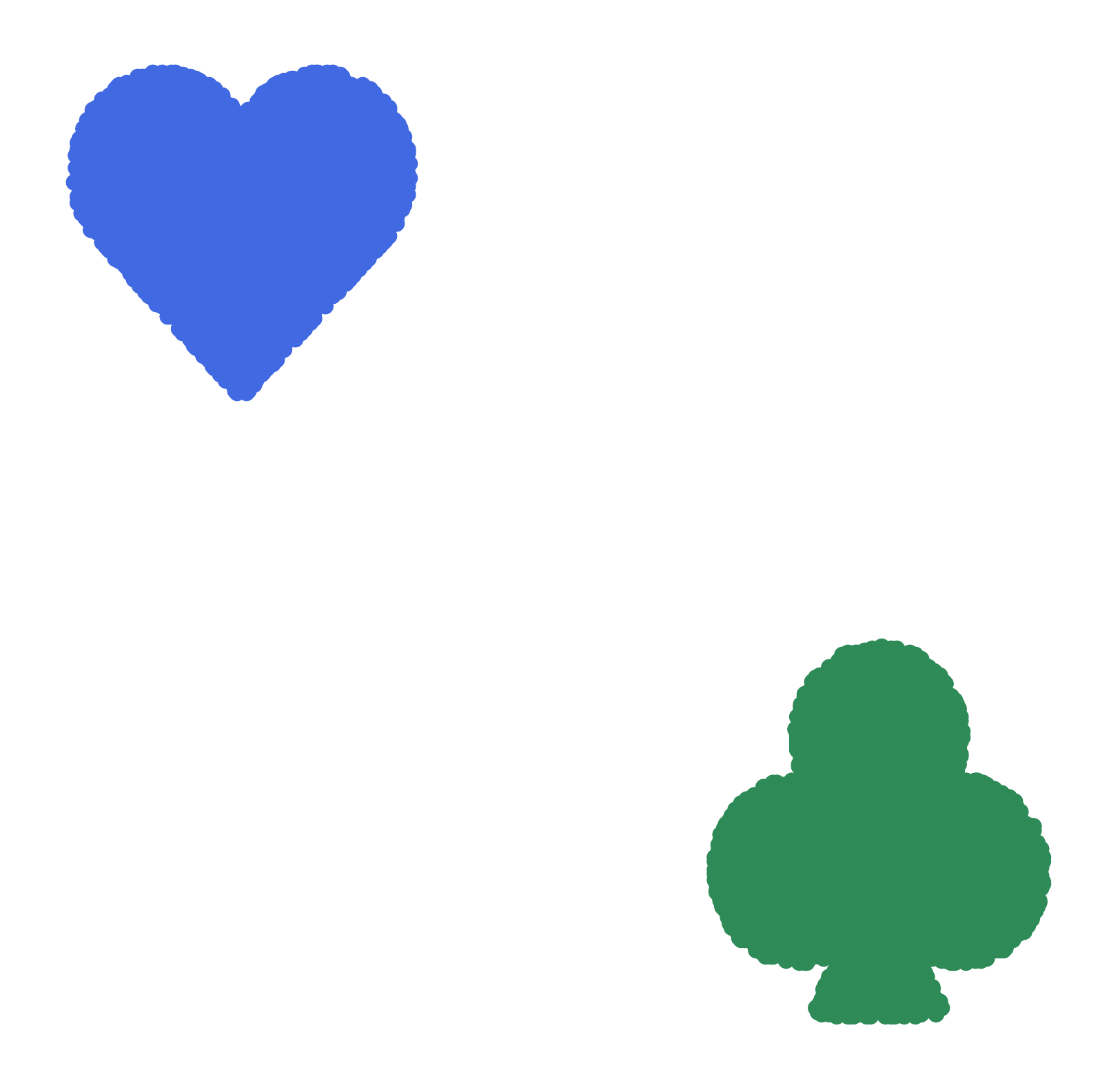}}
\hfill
\subfloat[
Fitting GMMs.
]{\includegraphics[width=0.21\linewidth,  trim={0mm 10mm 0mm 10mm}, clip]{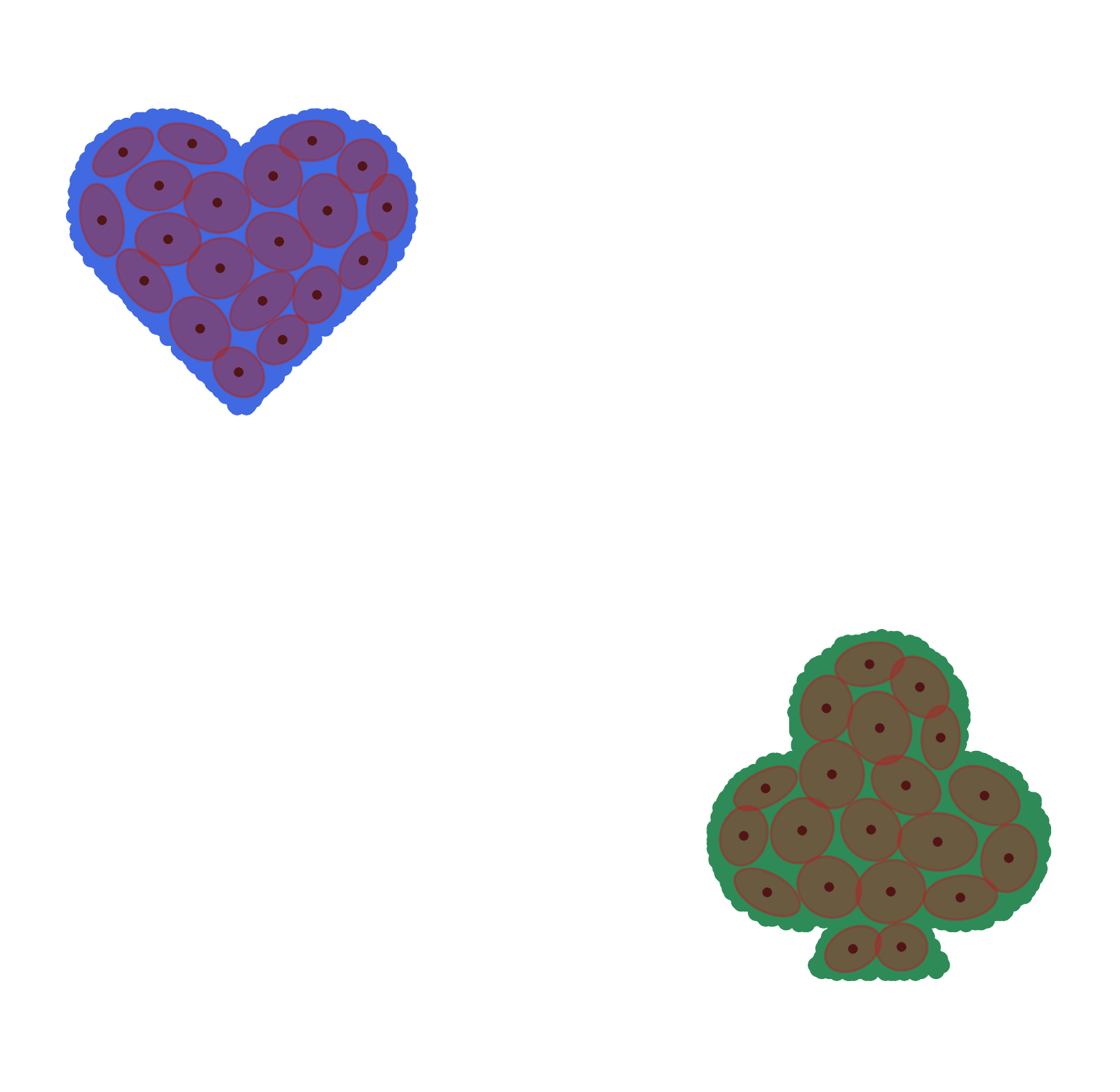}}
\hfill
\subfloat[
Matching means.
\label{fig:subplot_matching}
]{\includegraphics[width=0.2\linewidth,  trim={0mm 0mm 0mm 10mm}, clip]{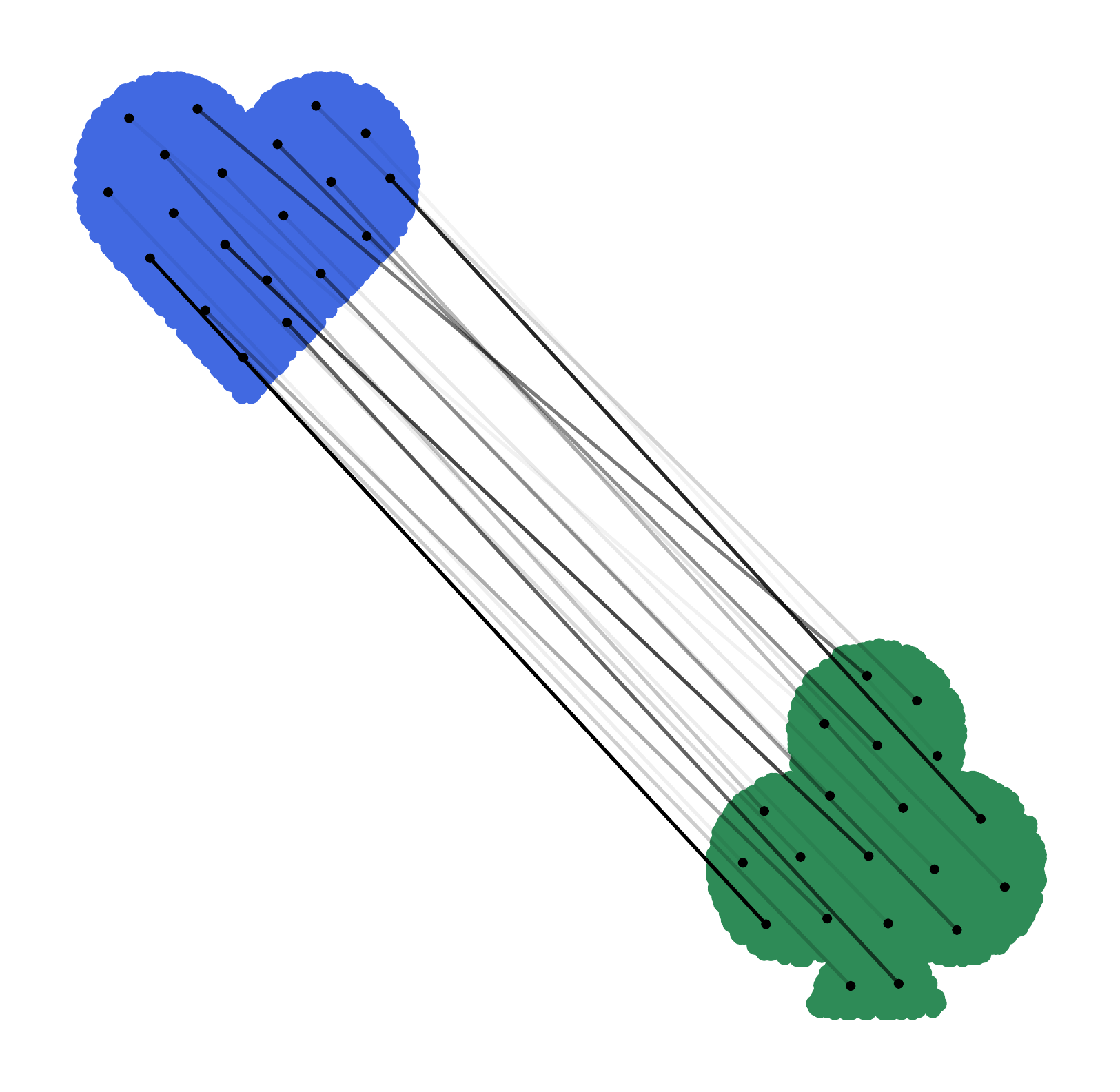}}
\hfill
\subfloat[
Partial transport.
]{\includegraphics[width=0.2\linewidth,  trim={0mm 0mm 0mm 10mm}, clip]{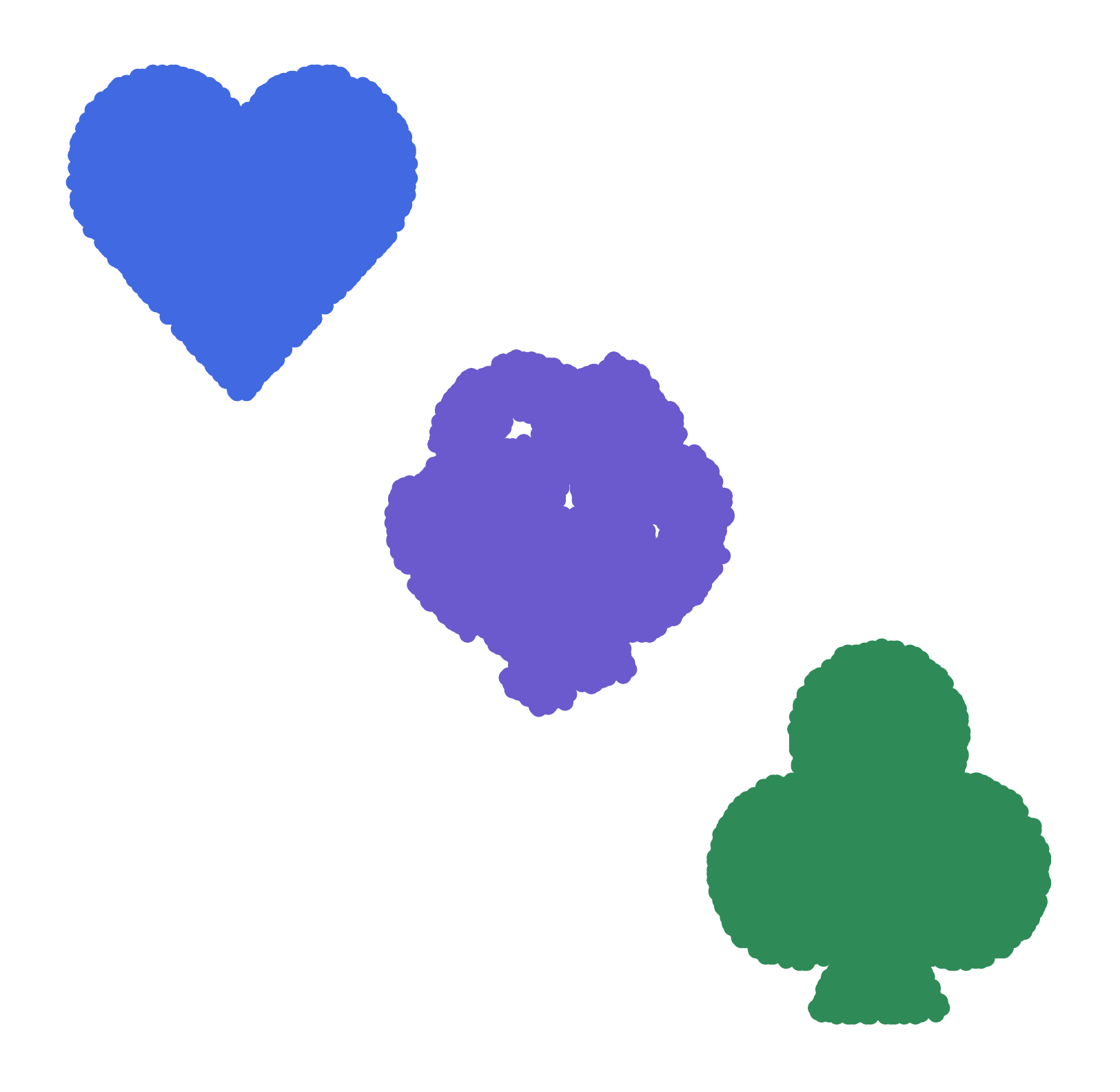}}
\vspace{-1mm}
\caption{\textbf{Illustration of the working principle of the approximation of the barycenter:}
For each group $g$ (e.g., heart and clover),  the algorithm receives a data matrix $\bX^g$.
It computes a GMM for each group.
It uses the means of these GMMs as an approximation of the geometry.
Next, it matches these means.
Finally, it computes the Bures-Wasserstein barycenter for each matched pair.
}
\vspace{-5mm}
\label{fig:ilustration}
\end{figure}

\paragraph{Approximate a barycenter.} 
%
		\begin{algorithm}[H]
			\small
			\caption{Approximate Barycenter}
			\label{alg:gmm_wasserstein_barycenter}
			\begin{algorithmic}[1]
				\Require{$\bX^g \in \reals^{|g|\times p}$ for $g \in \{a, b\}$, number $k$ of components, and weights $\blambda \in \Delta^2$.}
				
				\State Estimate parameters of L-BW: use Algorithm \ref{alg:gmm_wasserstein}
				
				\State Initialize: $h = 1$
				\For{$(i, j) \in [k]^2$}
				\If{$\bA_{i, j} = 1$}
				\State  $\bar{\bmean}_h  \leftarrow $  $\blambda_0\, \mathbf{m}_i^a + \blambda_1\, \mathbf{m}_j^b$.
				\State  $\bar{\bSigma}_h \leftarrow$ use Algorithm \ref{alg:gaussian_barycenter} with $ \bSigma_i^a$ and $\bSigma_j^b$.
				\State Set $h = h + 1$
				\EndIf
				\EndFor
				\State\Return{Parameters of barycenter $\{(\bar{\bmean}_h, \bar{\bSigma}_h)\}_{h=1}^{k}$.}
			\end{algorithmic}
		\end{algorithm}
%
Let $c$ be the barycenter of the distributions $a$ and $b$.
To approximate this barycenter, we use algorithm \ref{alg:gmm_wasserstein} 
to compute the parameters of the L-BW for $a$ and $b$.
Then, we calculate 
the Bures-Wasserstein barycenter of every $(i, j)$-pair that satisfies the match condition. 
%
Thus, the  approximated barycenter $c$ is in the space of $k$ Gaussian mixtures with
parameters $\{(\bar{\bmean}_h, \bar{\bSigma}_h)\}_{h=1}^{k}$.
%
%
The algorithm \ref{alg:gmm_wasserstein_barycenter} presents the pseudo-code
 of the approximate barycenter by L-BW\footnote{
 	Note that using one-vs-rest allows us to extend this method to more than two distributions.}.
Fig.\ref{fig:ilustration} shows an illustration of this procedure.

\section{Related work}
%
%
%
%
%
%
\paragraph{Mapping Estimation (ME).} 
%
%
%
To extend OT-based methods to unseen data, \cite{perrot2016mapping}
proposes an algorithm to learn the coupling and an approximation of the transport map.
%
This optimization problem learns a transformation regularized by a transport map.
However, ME relies on kernel-based methods, 
which usually have a time complexity of $O(n^3)$ \cite{bach2005predictive} for $n$ samples.
Additionally, it requires tuning the parameters associated with each kernel 
and the regularization parameter.
%
%
\paragraph{Practical computation of the barycenter.}
\cite{bonneel2011displacement} proposes a first approach 
to approximate the barycenter using signal decomposition 
and partial transport for multiple frequency bands.
%
%
%
%
%
%
However, it requires the nontrivial task of selecting the frequency bands.
%
A second approach by \cite{cuturi2014fast} proposes a fast algorithm to compute the free support Wasserstein barycenter.
This algorithm optimizes the locations of the barycenter and not the weights associated to the discrete measure.
It uses $k$ atoms or points to constrain the support of the barycenter. 
It is similar to our method, as both rely on the approximation of the geometry by $k$ atoms/components. 
%
However, our approach requires fewer atoms because the covariance matrices of the Gaussian mixtures encode more geometrical information.
\section{Applications \label{sec:applications}}
\vspace{-1mm}
%
We conduct a series of experiments to highlight the versatility
of Local Bures-Wasserstein (L-BW) transport.
We consider two tasks:  
\emph{i)} computing the barycenter of clouds of points
and \emph{ii)} building fair classifiers.
Both problems use an
approximation of the barycenter and the transport map.

We denote $\hat{\bT}_{a}^{c}(\cdot)$ the learned transport map from population $a$ to $c$, 
where $c$ is the learned barycenter of densities $a$ and $b$. 
We test the learned transport map on held-out data, 
in a cross-validation scheme splitting the data ten times at random.
Each time, we learn the barycenter and the transport map on $70\%$ 
of the data and apply it to the other $30\%$.
Besides L-BW,  we also investigate Mapping Estimation (ME) \cite{bonneel2011displacement} 
with both, linear and Gaussian kernels.
For the ME, we first compute the free support Wasserstein barycenter \cite{cuturi2014fast}, 
and then we use ME to learn the mapping to this barycenter.
%

Henceforth, we refer to $k$ as the number of atoms.
It represents, both, the number of components in L-BW and the number of points in the free
support Wasserstein barycenter used in ME.

\paragraph{Technical aspects.}
We used standard data processing and classification 
methods implemented in 
scikit-learn \cite{scikit-learn}.
We used the POT library \cite{flamary2017pot} for the 
convolutional Wasserstein barycenter and Mapping Estimation.
We used Scipy \cite{jones2014scipy} for the Hungarian algorithm. 
Our method is available\footnote{\href{https://github.com/ahoyosid/Local_Bures_Wasserstein}{Link to the Github repository}}.


\subsection{Shape interpolation of clouds of points}
A straightforward application of Wasserstein barycenter is the interpolation of shapes,
as it relies on performing optimal transportation over geometric domains.
We vary the number of atoms $k \in \{2, \ldots, 100\}$ and evaluate the performance 
of the approximated transport map 
and 
barycenter 
with two measures: 
\emph{i)} support recovery and
\emph{ii)} computation time.
For the ME, we set the regularization parameters of the linear and Gaussian kernels to $\eta = 10^{-3}$ and $\sigma = 1$, respectively\footnote{
	These are the parameters by default in the POT package \cite{flamary2017pot}.}.

\textbf{Synthetic dataset.}
We generated a synthetic dataset that consists of three binary $200\times 200$ images: 
cat, rabbit, and tooth 
(see Fig.6 in supp. materials). 
We put them  in a $1200 \times 1200$ square grid. 
We sample $\approx 10\, 000$ points at random inside every silhouette.
We aim to calculate the barycenter of these images given some weights 
$\blambda \in \Delta^3$.

\paragraph{Quality assessment.}
%
%
\begin{figure}[t]
	\small
	\centering
	\subfloat[Support recovery. \label{fig:support_recovery}]{\includegraphics[width=0.4\linewidth, trim={0 0mm 0 0mm}, clip]{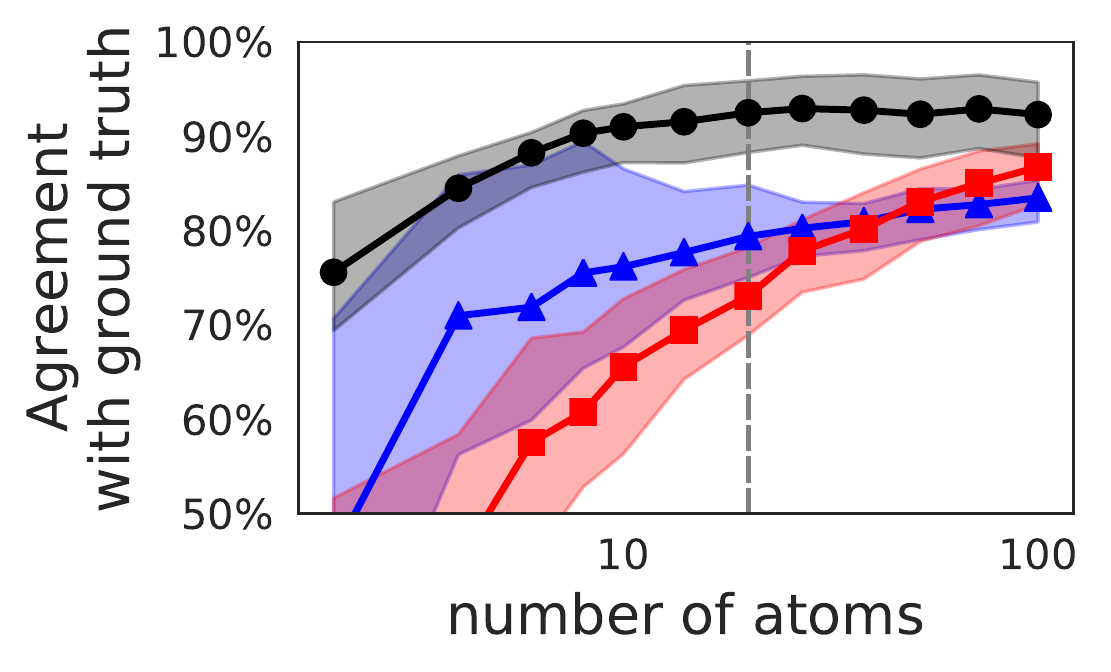}}
	%
	%
	\subfloat[Computation time. \label{fig:computation_time}]{\includegraphics[width=0.4\linewidth,  trim={0 0mm 0 0mm}, clip]{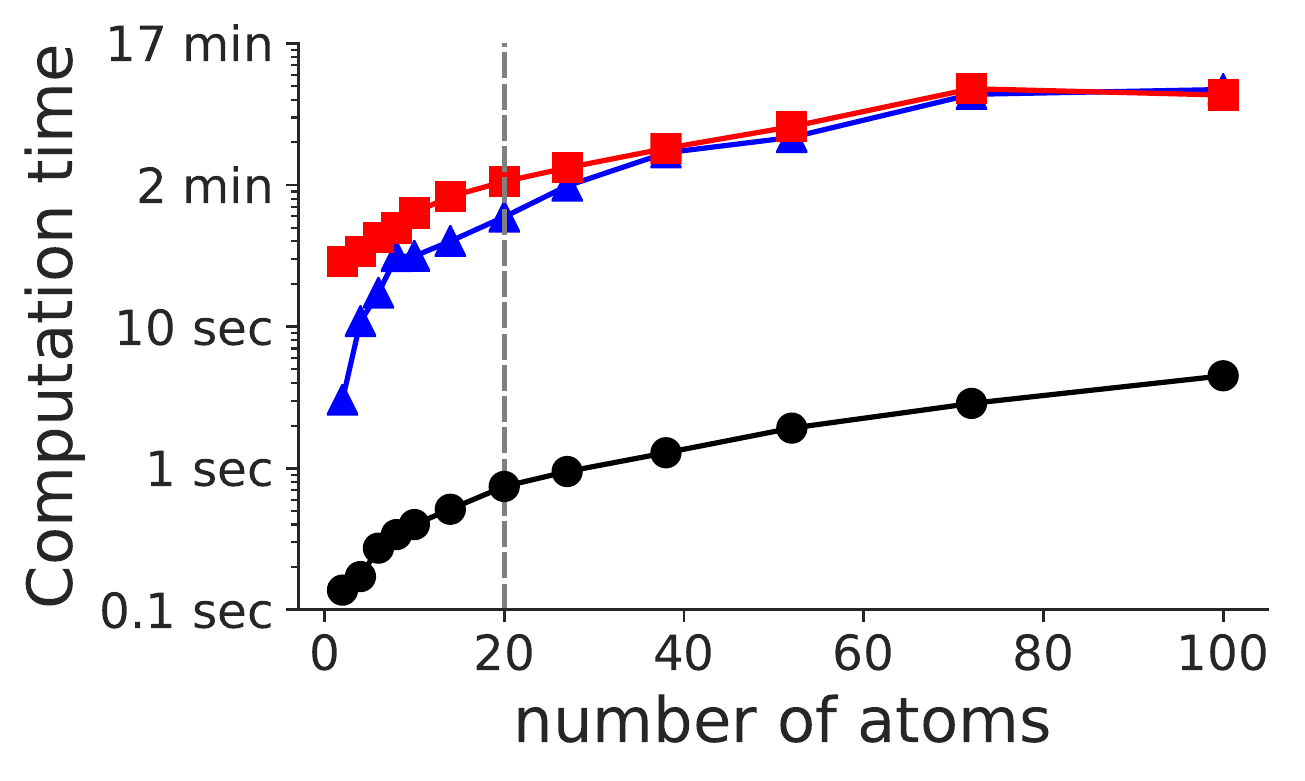}}\\
	\subfloat{\includegraphics[width=0.4\linewidth, trim={0 20mm 0 20mm}, clip]{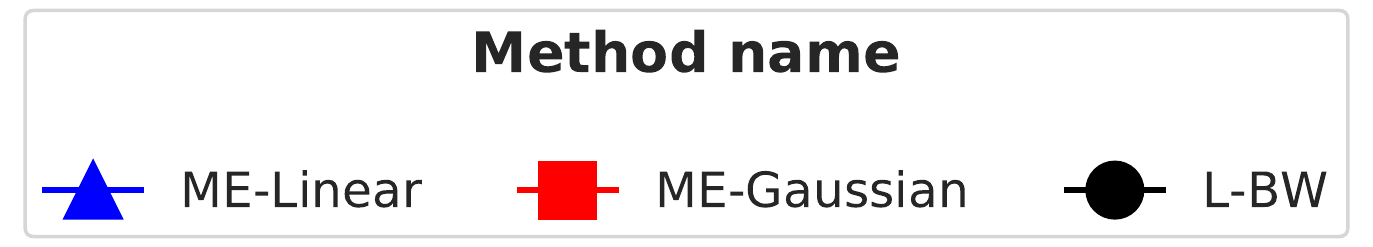}}	
	\caption{
		\small
		\textbf{Performance evaluation.}
		The vertical dashed line denotes $k=20$.
		%
		L-BW requires fewer atoms to recover the support, and it is the fastest method.
	}
	\label{fig:barycenter_computation_time}
\end{figure}
To compare the quality of the approximated barycenter, 
we measure its agreement proportion with ground-truth. 
The ground-truth is in itself a silhouette, and we build it using 
the convolutional 2-D Wasserstein barycenter algorithm  \cite{solomon2015convolutional}\footnote{
	To use this algorithm, the images are normalized to sum to $1$ and 
	concatenated in a third dimension.}, 
followed by binarization by thresholding. 
We set the threshold via the Otsu's method \cite{otsu1979threshold}.

\begin{figure}[t]
	\centering
	\small
	\vspace{-3mm}
	\subfloat[\textbf{Baseline:}  $2$-D convolutional Wasserstein\cite{solomon2015convolutional}. \label{fig:2d_barycenters_baseline}]{
		\includegraphics[width=0.21\linewidth, trim={25mm 20mm 20mm 20mm}, clip]{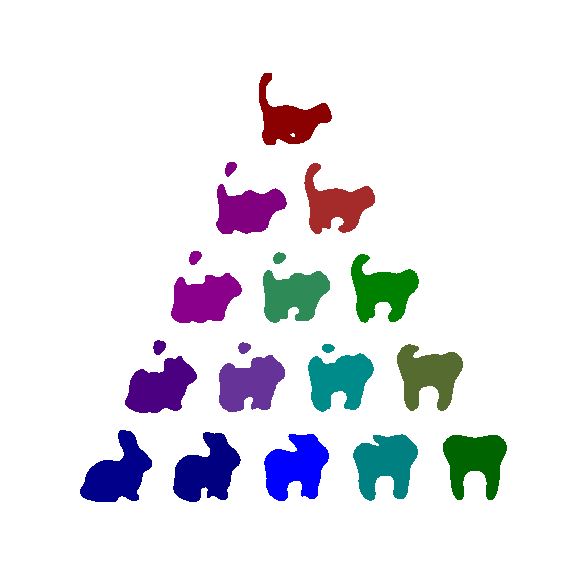}
	}
	\hfill
	\subfloat[ME-Linear: \newline $20$ atoms. ]{ 
	\includegraphics[width=0.21\linewidth, trim={30mm 20mm 20mm 20mm}, clip]{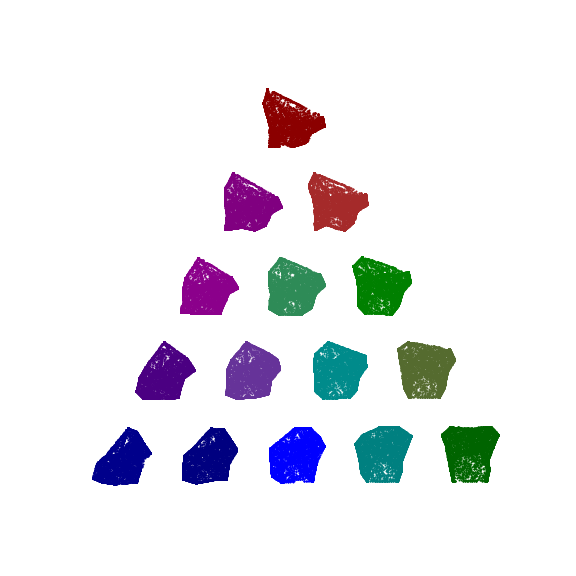}
	}
	\hfill
	\subfloat[ME-Gaussian: \newline $20$ atoms. ]{ 
	\includegraphics[width=0.21\linewidth, trim={30mm 20mm 20mm 20mm}, clip]{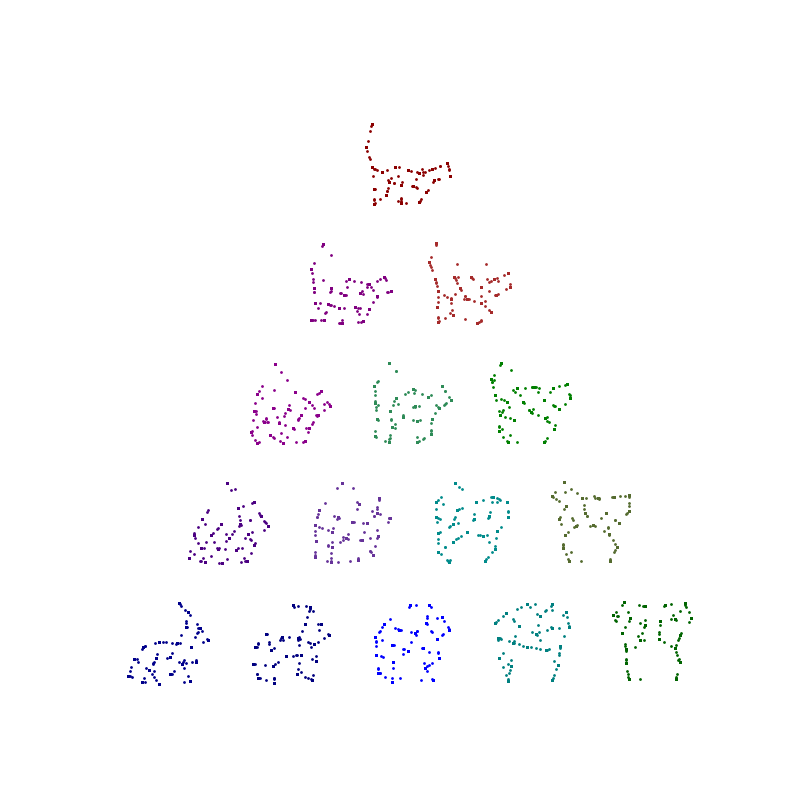}
	}%
	\hfill
	\subfloat[L-BW: $20$ atoms. \label{fig:2d_barycenters_20}]{ 
		\includegraphics[width=0.21\linewidth, trim={30mm 20mm 20mm 20mm}, clip]{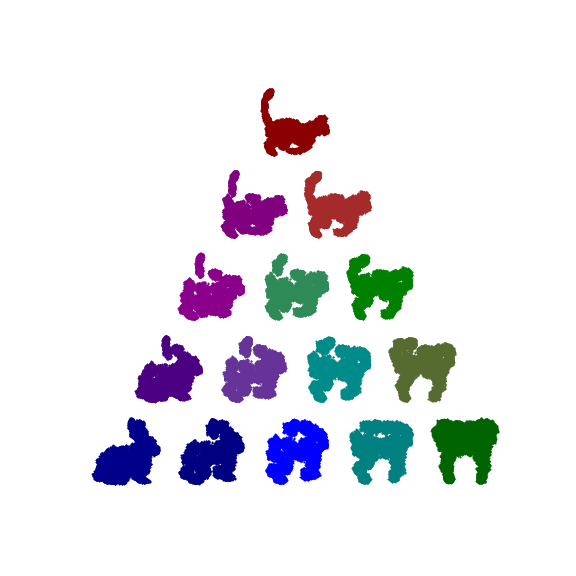}
	}%
	\vspace{-4mm}
	\caption{
		\textbf{Visualization of barycenters:} 
		The coordinates represent the Wasserstein simplex, which consists in all of their 
		Wasserstein barycenters under varying weights $\blambda \in \Delta^{3}$.
		Fig.\ref{fig:2d_barycenters_baseline} displays the barycenters obtained 
		by \cite{solomon2015convolutional} after thresholding using Otsu's method.
		%
		%
	    \vspace{-4mm}
		\label{fig:2d_barycenters}
	}
\end{figure}
Fig.\ref{fig:support_recovery} displays the support recovery of intermediate silhouettes
as a function of the number of atoms.
%
%
ME-based methods require five times more atoms than L-BW to improve the quality of their 
resulting images.
%
L-BW has higher percentage of agreement with ground truth 
than ME with both linear and Gaussian kernels.
It seems to reach a plateau at $k = 20$, which corresponds to
the best representation according to the Akaike information criterion 
(see Fig.7 in supp. materials).

Fig.\ref{fig:2d_barycenters} shows the barycenters found by various algorithms for $k=20$ on 
the synthetic dataset.
Fig.\ref{fig:2d_barycenters_baseline} displays our ground-truth. 
%
%
ME-Gaussian generates barycenters that reproduce the shapes, 
yet it concentrates the transported samples in a few atoms. 
In contrast, ME-Linear does not concentrate the transported samples; 
however, it fails to preserve the silhouette.
L-BW reproduces the transitions between inputs, 
and creates plausible intermediate silhouettes. 
   
\paragraph{Computation time.}
We measured the running time of various approaches, 
including both the training time and the transport time. 

Fig.\ref{fig:computation_time} gives computation time for different methods.
L-BW displays overall the lowest computation time for all atoms;
on average $80$ times faster than ME-Linear.
For $k < 40$, ME-Linear is ten times faster than ME-Gaussian; 
when the number of atoms increases, this factor drops to approximately 1.5.

\subsection{Demographic-parity fairness}
Another application of OT is to build fair classifiers.
We rely on demographic-parity,
%
which requires that predictions of an estimator cannot reveal 
information about the protected class any better (up to a tolerance) 
than random guessing \cite{olfat2017spectral, zafar2017fairness}. 

%
We consider mixing distributions of two groups (e.g., Female and Male) 
via OT to produce naive fair classifiers. 
We modify the input data $\bX^g$ by transporting each sample $\bsx$ to 
the barycenter $c$ of the two populations, $g \in \{a, b\}$. 
This is proposed in \cite{del2018obtaining} for Gaussian random variables.
We extend this idea to more complex distributions via L-BW,  as follows
\begin{equation}
D\left [\Probability[\mathbf{y} |\, \hat{\bT}_a^c(\bX^a);\, \theta],  
\Probability[\mathbf{y} |\, \hat{\bT}_b^c(\bX^b);\, \theta]\right] \leq \epsilon,
\end{equation}
where $\theta$ denotes the parameter of the classifier, $\mathbf{y}$ is the binary classification target, 
and $D(\mu, \nu)$ is a similarity measure of two distributions $\mu$ and $\nu$ 
(e.g., Kullback-Lieber, or Wasserstein).
$\epsilon$ is the desired tolerance.
%
This transformation of the input data is oblivious to the target; 
hence we do not have precise control of the tolerance $\epsilon$.

\paragraph{Fairness experiment.}
We present an empirical study on five real-world datasets.
In all classification tasks, we collect statistics concerning prediction 
and fairness score measure.
%
%
%
We use nested cross-validation 
to set hyperparameters. 
Our preprocessing pipeline comprises standardization and dimensionality 
reduction via Principal Component Analysis (PCA). 
We use one-hot-encoding to handle categorical variables.

%
%
We use the Demographic-Parity-$\gamma$ score \cite{olfat2017spectral} to measure fairness.
It is the maximum distance of the Receiver Operating Characteristic (ROC)
curve for the group membership label from the diagonal, 
where zero represents an absolute lack of predictability 
(see Fig.10 in supp. materials). 
Thus, the lower the fairness score, the better.
%
%
For prediction performance, we use the area under the ROC curve (AUC). 
This score upper bounded by $1$ and higher values mean better predictions.

\paragraph{Setting hyperparameters.} 
We vary the number of principal components $d \in \{2, 4, \ldots, 30\}$, 
and the number of atoms $k \in \{1, \ldots, 10\}$ for the L-BW.
We set the parameter to regularize the covariance of the GMMs to $10^{-6}$.
To compute the free support Wasserstein barycenter, 
we set the grid $k \in \{10, 20, 50, 100, 200, 300\}$.
We set grid for the parameters of ME to $\eta \in \{10^3, 10^2, 10^1, 1, 10\}$ 
and $\sigma \in \{1, 10, 10^2, 10^3\}$ for the linear and Gaussian kernels, respectively.
We set weights of the barycenter to $\blambda = [0.5, 0.5]$, 
the new distribution should not benefit any of two populations.

\paragraph{Datasets.}
\begin{wrapfigure}{R}{0.45\textwidth}
	\vspace{-8mm} 
	\centering
	\subfloat{\includegraphics[width=1.05\linewidth, trim={5 0mm 0mm 0mm}, clip]{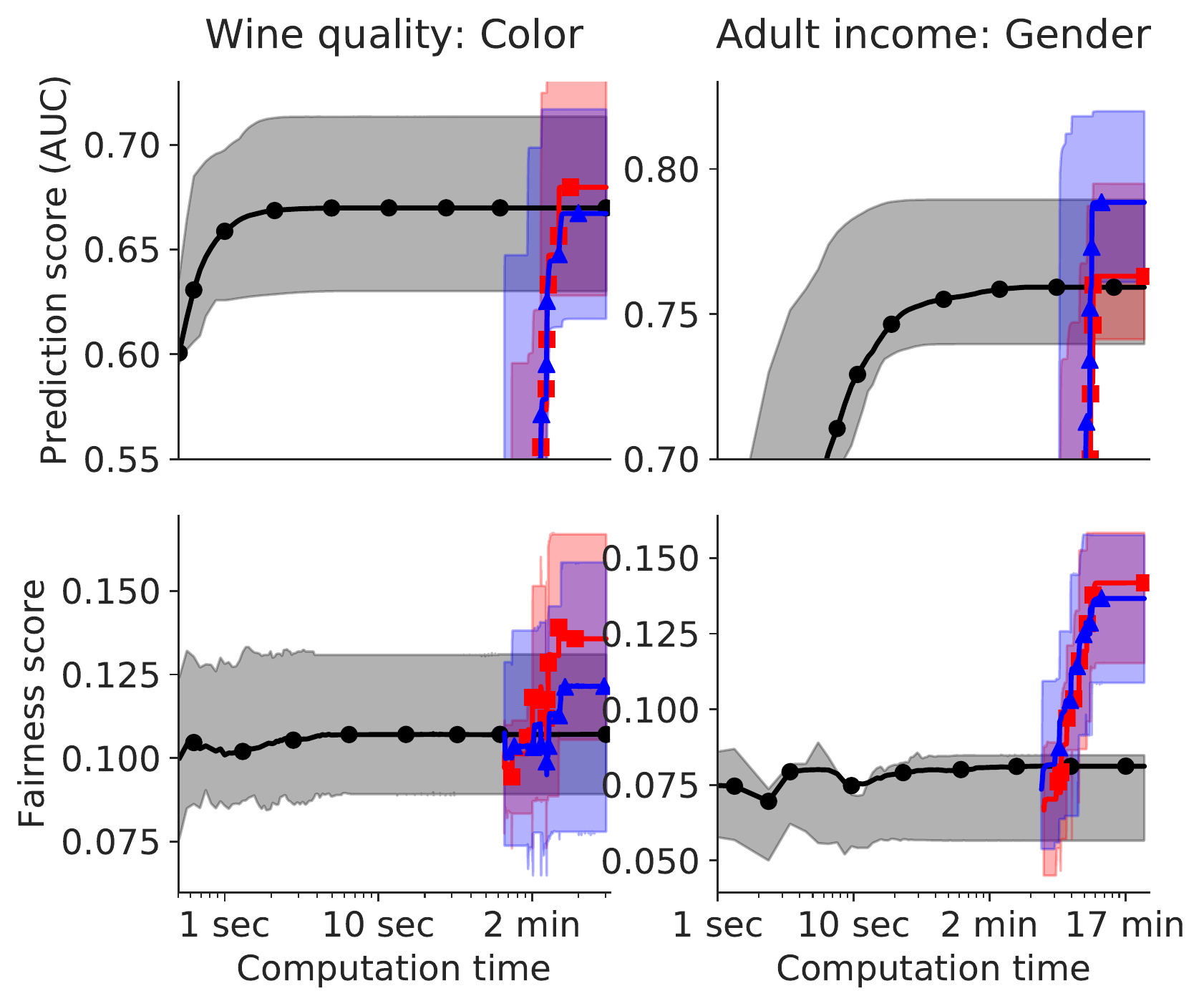}}\\
	\vspace{-3mm} 
	\subfloat{\includegraphics[width=.9\linewidth, trim={0 20mm 0 20mm}, clip]{legend_computation_time.pdf}}	
	\caption{\textbf{Computation time taken to reach a solution:} 
		Quality of the fit of a $\ell_2$ penalized logistic regression as a function 
		of the computation time for a fixed number of atoms.
		\emph{(top)} Predictive accuracy 
		and \emph{(bottom)} fairness score on held-out-data.
		In both datasets, L-BW obtains the best fairness score, 
		and the same prediction score than ME-Gaussian with less computation time to reach a stable solution. 
		The time displayed includes the computation of the barycenter and transport mapping.
	}
	\label{fig:time_to_reach_solution}
\end{wrapfigure}
We investigate our approach in several  binary  classification problems based on four datasets from the UCI Machine Learning Repository\cite{blake1998uci}.
%
%
See Fig.11 in supp. materials a visualization
of the first two principal components of these datasets conditioned on the sensitive groups.

\emph{Wine quality\footnote{\href{https://archive.ics.uci.edu/ml/datasets/wine+quality}{https://archive.ics.uci.edu/ml/datasets/wine+quality}}:} 
%
it contains $12$ attributes of $4\,898$ wines. 
%
Professional taste-tester provided a ranking out of $10$. 
%
The task is to predict if a wine has a rating higher or equal than $6$.
The protected groups are color-based (White and Red).
All explanatory variables are continuous. 

\emph{German Credit Data\footnote{
		\href{http://archive.ics.uci.edu/ml/datasets/statlog+(german+credit+data)}{http://archive.ics.uci.edu/ml/datasets/statlog+(german+credit+data)}}:}
%
it contains $20$ features ($7$ numerical and $13$ categorical) 
related to the economic situation of $1\,000$ German applicants for loans. 
%
The aim is to predict whether or not an applicant is going to default the credit loan.
The protected groups are gender-based (Female and Male).

\emph{Adult Income\footnote{\href{https://archive.ics.uci.edu/ml/datasets/adult}{https://archive.ics.uci.edu/ml/datasets/adult}}:}
%
it contains $14$ features concerning demographic 
characteristics of $45\,222$ individuals. 
The task is to predict whether or not a person has more
than $50\,000$ as an income per year. 
The protected groups are education-based (Higher and Lower education)
 and gender-based (Male and Female). 

\emph{Taiwan credit\footnote{\href{https://archive.ics.uci.edu/ml/datasets/default+of+credit+card+clients}{https://archive.ics.uci.edu/ml/datasets/default+of+credit+card+clients}}:}
%
it contains the credit information of $30\,000$ individuals 
%
$24$  features represent this information, where $9$ of them are categorical. 
The task is to predict customers if default payments.

\emph{COMPAS 
	 \cite{larson2016we}:} 
it is a collection of $10\, 000$ criminal defendants screened in Broward County, Florida U.S. in $2013$ and $2014$.
The features are demographics and criminal records of offenders. 
The task is to score an individual's likelihood of reoffending. 
%
%
The protected groups are race-based (Black and White) and gender-based (Male and Female).

\paragraph{Fairness on a time budget.}
\begin{wrapfigure}{R}{0.5\textwidth}
	\centering
	\includegraphics[width=0.4\textwidth, trim={0 3mm 0 0}, clip]{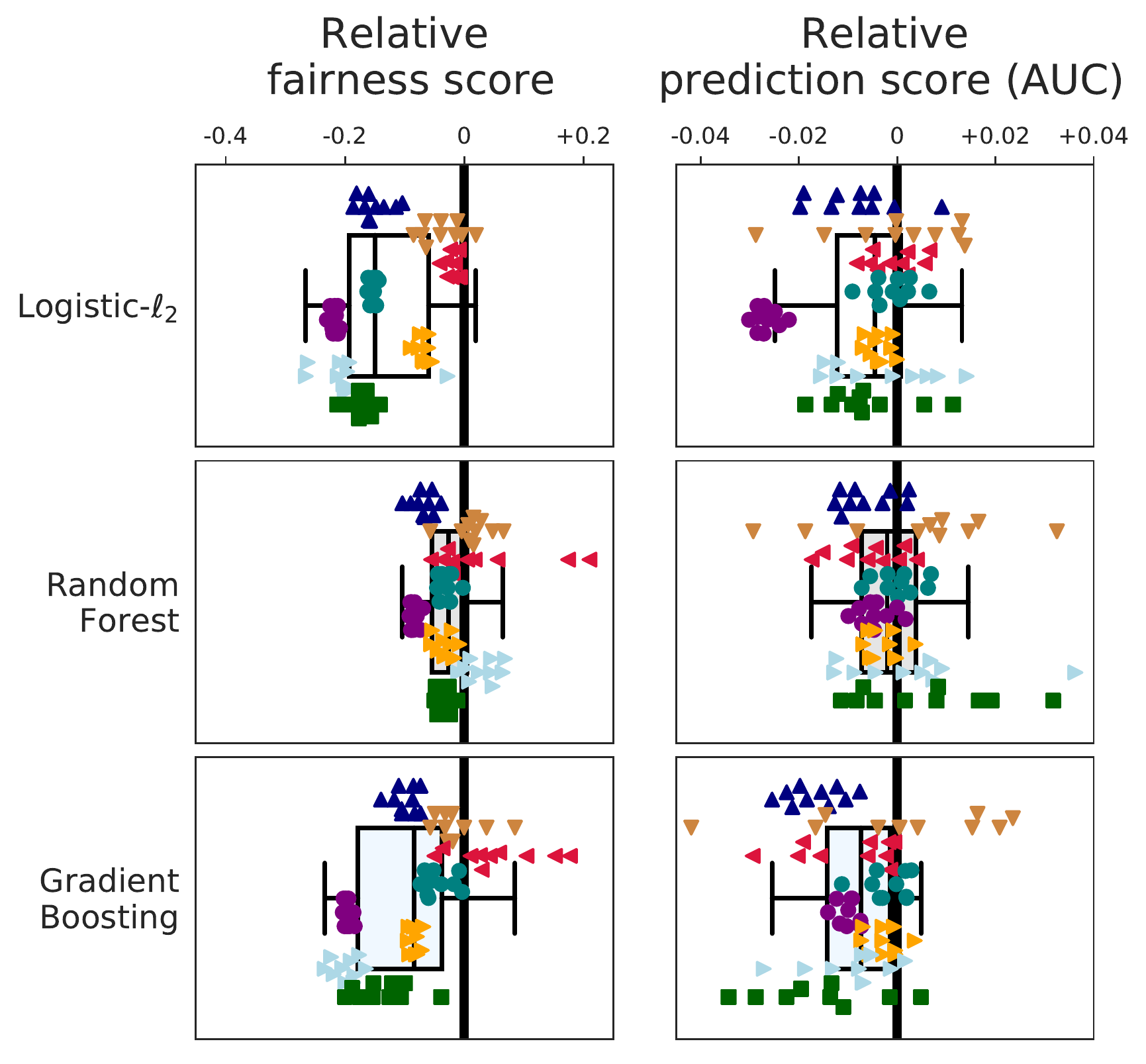}\\
	\hfill
	\includegraphics[width=0.9\linewidth]{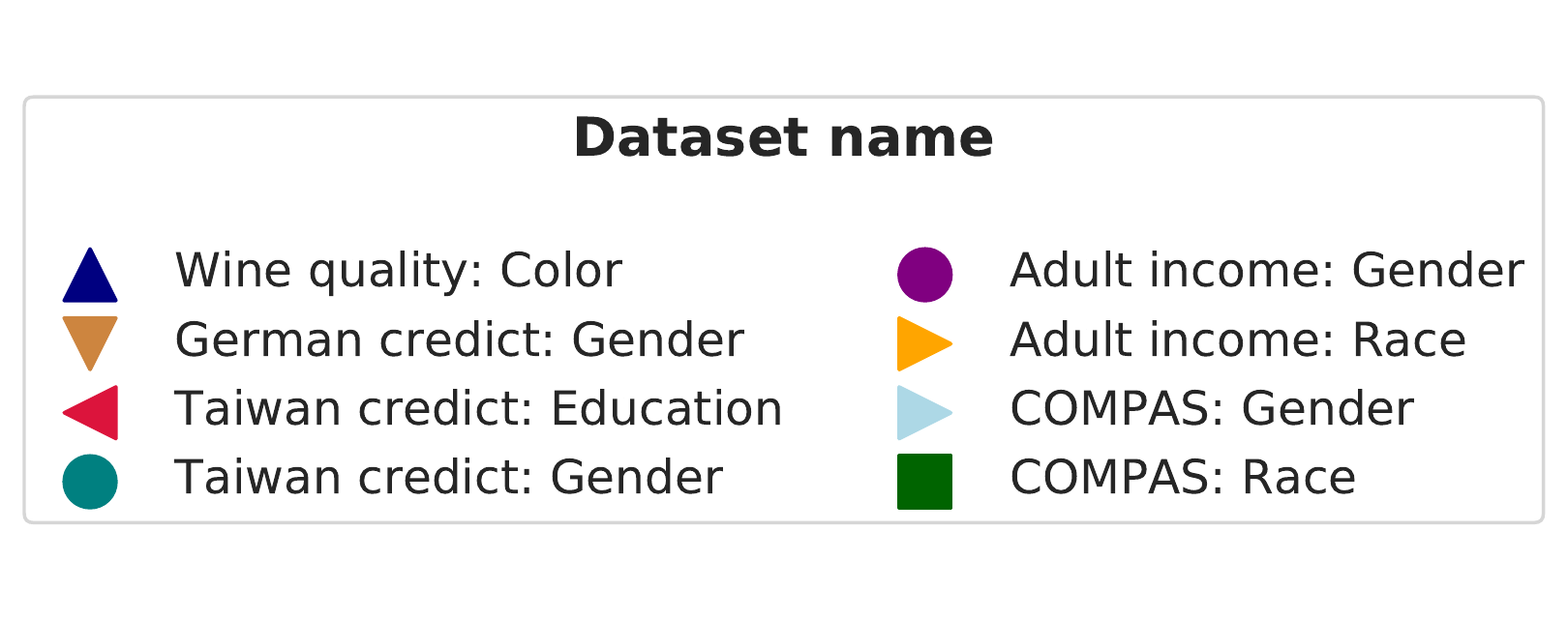}
	\vspace{-3mm}
	\captionof{figure}{\textbf{Impact of mixing distributions with L-BW on performance of various classifiers:} 
		The performance of classifiers after mixing sensitive groups relative to 
		mean scores of the raw classifier (per dataset). 
		%
		%
		The fairness score, the lower, the better ($<0$).
		The AUC, the higher, the better ($>0$). 
		L-BW improves the fairness of logistic regression while maintaining prediction accuracy.
		For Gradient Bosting, L-BW reduces the fairness score at the price of decreasing accuracy.
		The performance of Random Forest remains the same.
	}
	\label{fig:relative_performance_plugin}
\vspace{-8mm}
\end{wrapfigure}
We examine the impact of mixing groups on prediction time for various approximate transport map methods.  
%
%
We are interested in the total computation time needed to learn a model: the cost of computing the
approximate representation (i.e., fitting a barycenter and a transport mapping) and of training the classifier.

Fig.\ref{fig:time_to_reach_solution} shows the prediction accuracy as a 
function of the computation time for training 
a logistic regression with $\ell_2$-penalty, and
it displays the mean the  across folds and the percentiles $25$ and $75$.
In both datasets, L-BW requires less time to converge to an estimator with a lower fairness score and 
displays the same predictive performance as ME-Gaussian.
ME-Linear has the best classification performance in the income dataset; 
however, it has a more significant unfair behavior than  L-BW.

\paragraph{Fair classifiers.}
We analyze the impact of mixing groups by L-BW in 
various conventional classifiers: 
Logistic regression ($\ell_2$ penalty), 
Gradient Boosting Trees ($100$ trees), and 
Random Forest ($100$ trees). 
See Fig.13 in supp. materials for more classifiers.

Fig.\ref{fig:relative_performance_plugin} shows the performance classifiers after 
mixing groups with L-BW.
The performance is relative to the mean performance of the raw classifier (per dataset). 
The fairness score (DP-$\gamma$) of Random-Forest does not display any improvement, 
whereas the other classifiers present significantly better results ($\text{p}_\text{value} < 10^{-8}$, 
paired Wilcoxon rank test).
L-BW decreases the predictive performance of Gradient Boosting Trees to the raw classifier. 
This drop in accuracy is not significant for other classifiers.

\section{Discussion}
\vspace{-1mm}
We proposed a fast method to compute an approximation of the barycenter and 
transport mapping, which can be used in out-of-sample data.
Our method directly exploits highly concentrated samples in low 
dimensions to approximate the Wasserstein barycenter 
as well as the transport map.
First, it approximates the training data with Gaussian mixtures. 
For each distribution, we fix the means of these Gaussians as the new geometrical reference.
Then, we select the closest means between densities.
Finally, we perform pointwise Gaussian transport between matched components.
%
%
%
Additionally, we propose an extension of our approach to more than two densities, 
which has linear complexity in the number of densities. 
From a machine learning perspective, 
we can directly project held-out-data into the approximate barycenter, 
which makes it possible to use in standard machine learning pipelines.

We demonstrated the effectiveness of our approximation on two problems:
shape interpolation of clouds of points and naive demographic-parity fair classifiers.
In both problems, our method displays the fastest computation time,  outperforming the ME.
L-BW extends \cite{del2018obtaining} to densities more complex than Gaussians.
It  requires fewer components to approximate the barycenter and, for most classifiers,
 it produces naive fair estimators while maintaining prediction accuracy.
Our approach is similar to \cite{cuturi2014fast}, as it relies on $k$ Gaussian components (atoms).
Yet, our approach also takes the covariance of subpopulations into account. 
Experimental results confirm that this additional information benefits the support recovery of the barycenter.



However, the current framework has the following limitations: 
\emph{i)}  it assumes that the few clusters represent the densities involved, 
which limits its applications to low/medium dimensionality settings;
\emph{ii)}  it assumes that the number of atoms required to approximate 
each distribution included in the barycenter calculation is the same,
which can be problematic for some data;
\emph{iii)} the theoretical properties of the L-BW are out of the scope of this paper.
Addressing the above limitations are new directions for future work.


\medskip
\small

\bibliographystyle{abbrvnat}
\bibliography{biblio}

\section{Applications: Additional analysis}
\subsection{Shape interpolation of a cloud of points experiment \label{sec:supp_0}}
%
\noindent
\begin{minipage}[t]{\textwidth}
		\begin{minipage}[t]{0.47\textwidth}
		\vspace{10mm}
		\centering\raisebox{\dimexpr \topskip-\height}{%
			\includegraphics[width=0.9\linewidth]{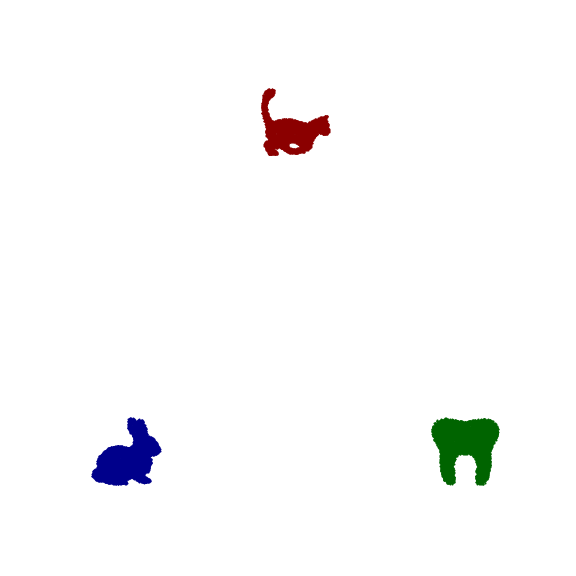}}
		\captionof{figure}{\textbf{Input shapes:}  
			Cat, rabbit, and tooth.
			Each silhouette contains $\approx 10\, 000$ points.} 
		\label{fig:shapes}
	\end{minipage}
	\hfill
	\begin{minipage}[t]{0.47\textwidth}
		\vspace{10mm}
		\centering\raisebox{\dimexpr \topskip-\height}{%
				\includegraphics[width=0.9\linewidth]{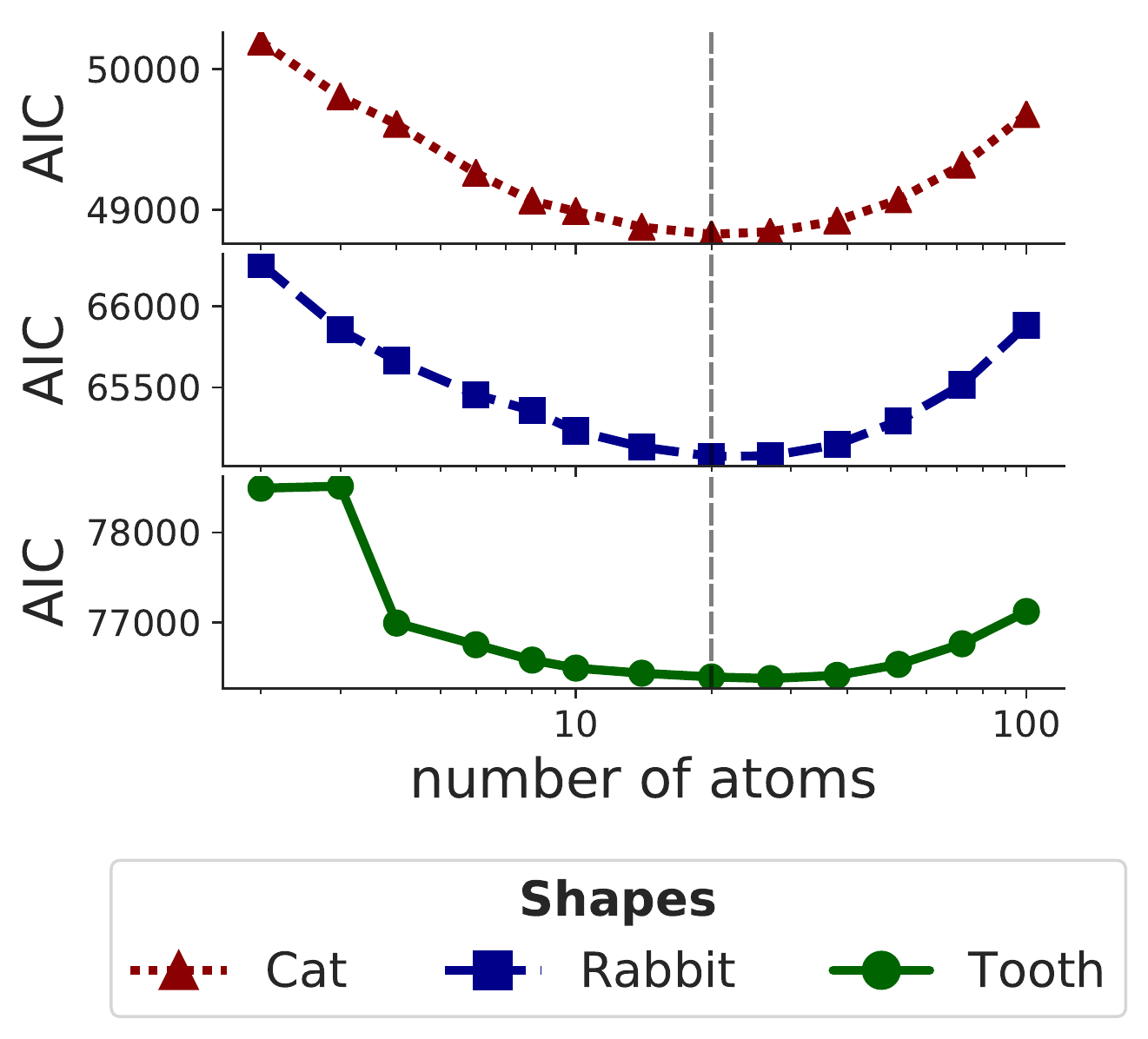}}
	\captionof{figure}{\textbf{Setting the number of components:}  The Akaike information criterion (AIC) as a function of the number of atoms for each silhouette.
	$k = 20$ achieves the minimum AIC for all shapes.  
	}
	\label{fig:2D_barycenter_parameters}
	\end{minipage}
\end{minipage}
\begin{figure}[H]
	\centering
	%
	\subfloat[ME (linear): $10$ atoms.]{ 
		\includegraphics[width=0.32\linewidth, trim={20mm 25mm 20mm 25mm}, clip]{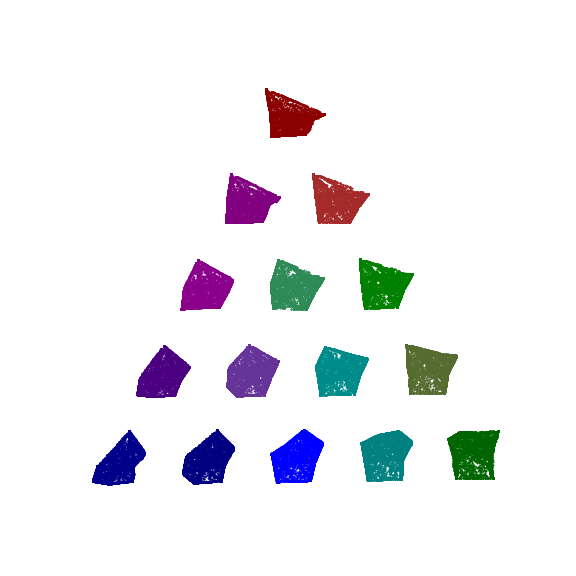}
	}%
	\subfloat[ME (Gaussian): $10$ atoms.]{ 
		\includegraphics[width=0.32\linewidth, trim={20mm 25mm 20mm 25mm}, clip]{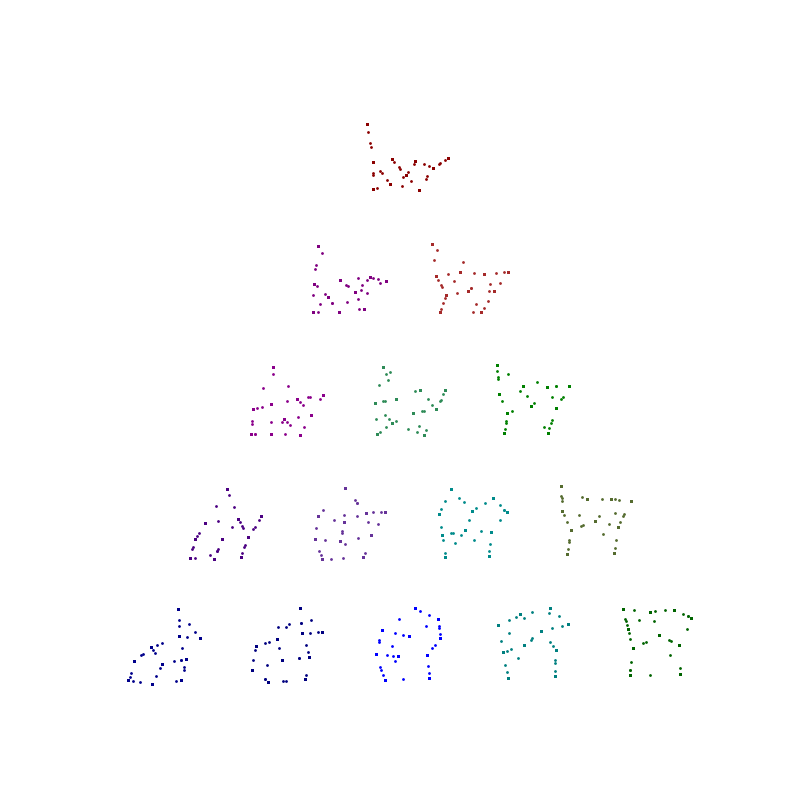}
	}%
	%
	%
	\subfloat[L-BW: $10$ atoms. \label{fig:2d_barycenters_10}]{
		\includegraphics[width=0.32\linewidth, trim={20mm 25mm 20mm 25mm}, clip]{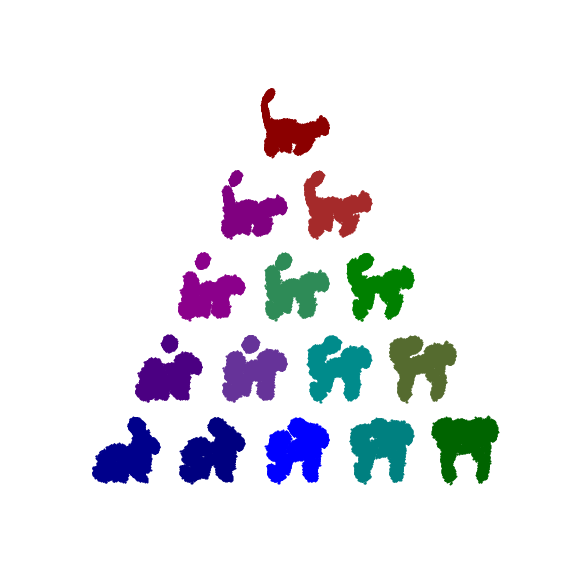}
	}\\%
	\subfloat[ME (linear): $100$ atoms.]{ 
		\includegraphics[width=0.32\linewidth, trim={20mm 25mm 20mm 25mm}, clip]{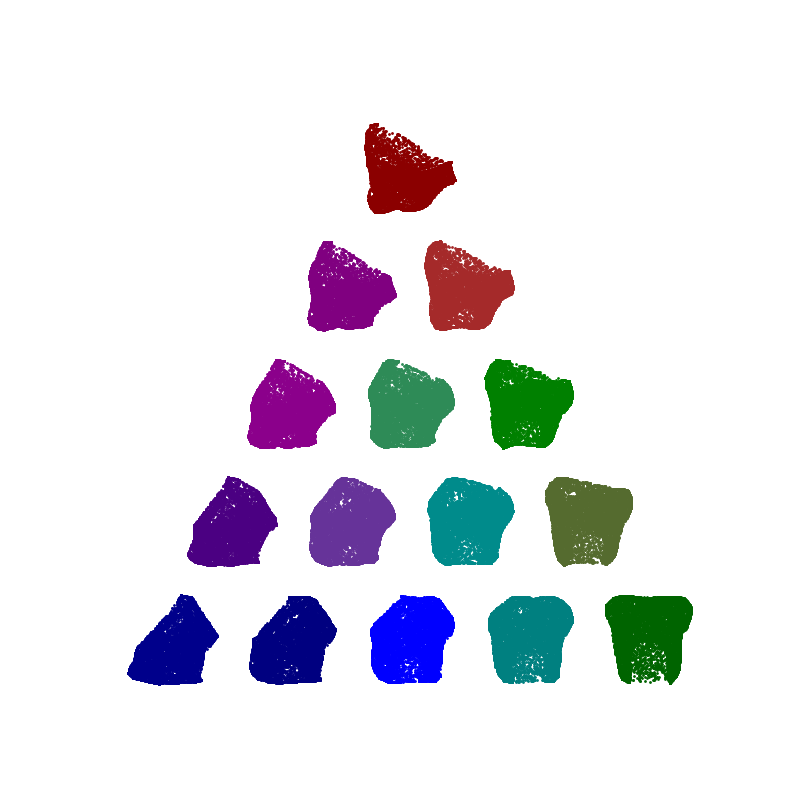}
	}%
	\subfloat[ME (Gaussian): $100$ atoms.]{ 
		\includegraphics[width=0.32\linewidth, trim={20mm 25mm 20mm 25mm}, clip]{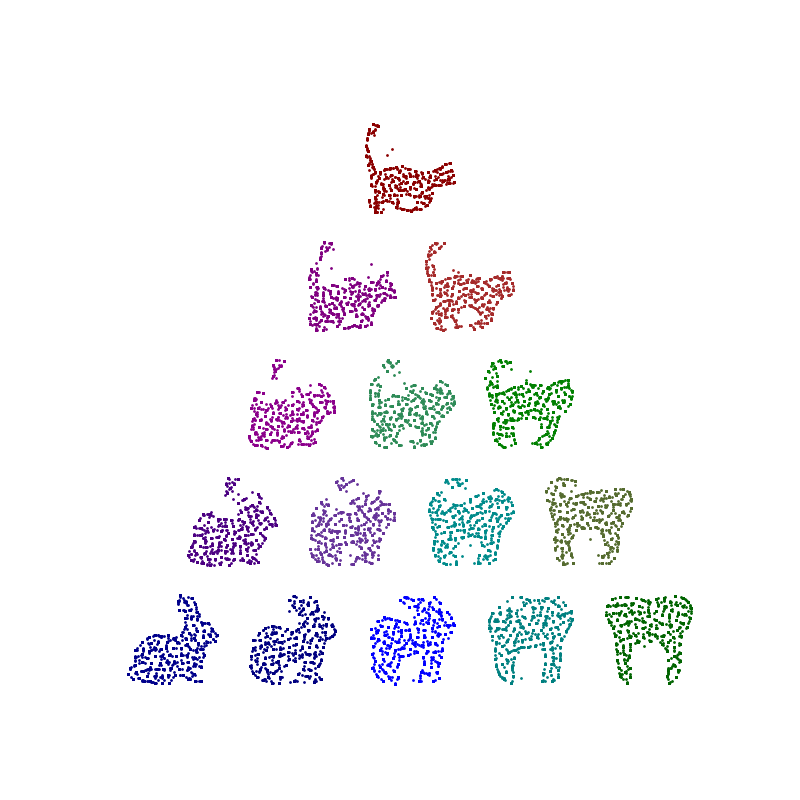}
	}%
	\subfloat[L-BW: $100$ atoms. \label{fig:2d_barycenters_100x}]{ 
		\includegraphics[width=0.32\linewidth, trim={20mm 25mm 20mm 25mm}, clip]{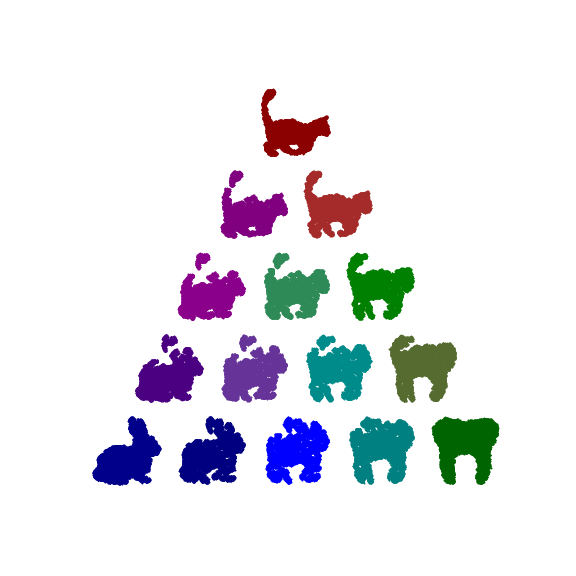}
	}%
	\caption{
		\textbf{Visualization of barycenters for different methods:} 
		The coordinates represent the Wasserstein simplex, which consists in all of their 
		Wasserstein barycenters under varying weights $\blambda \in \Delta^{3}$.
		\emph{(top)} row corresponds to $k=10$;
		\emph{(bottom)} row corresponds to $k=100$. 
		\vspace{-4mm} \label{fig:2d_barycenters_supp}
	}
\end{figure}

\begin{figure}
	\centering
	\small
	\subfloat[ME (Linear)]{\includegraphics[width=0.95\linewidth]{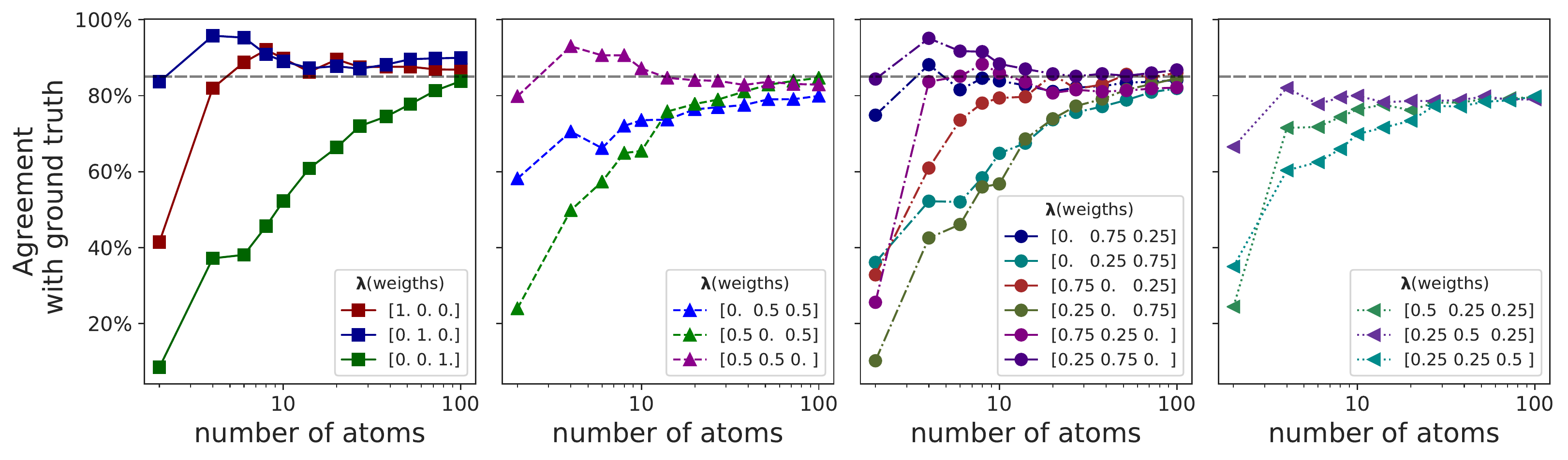}}\\
	\vspace{-3mm} 
	\subfloat[ME (Gaussian)]{\includegraphics[width=0.95\linewidth]{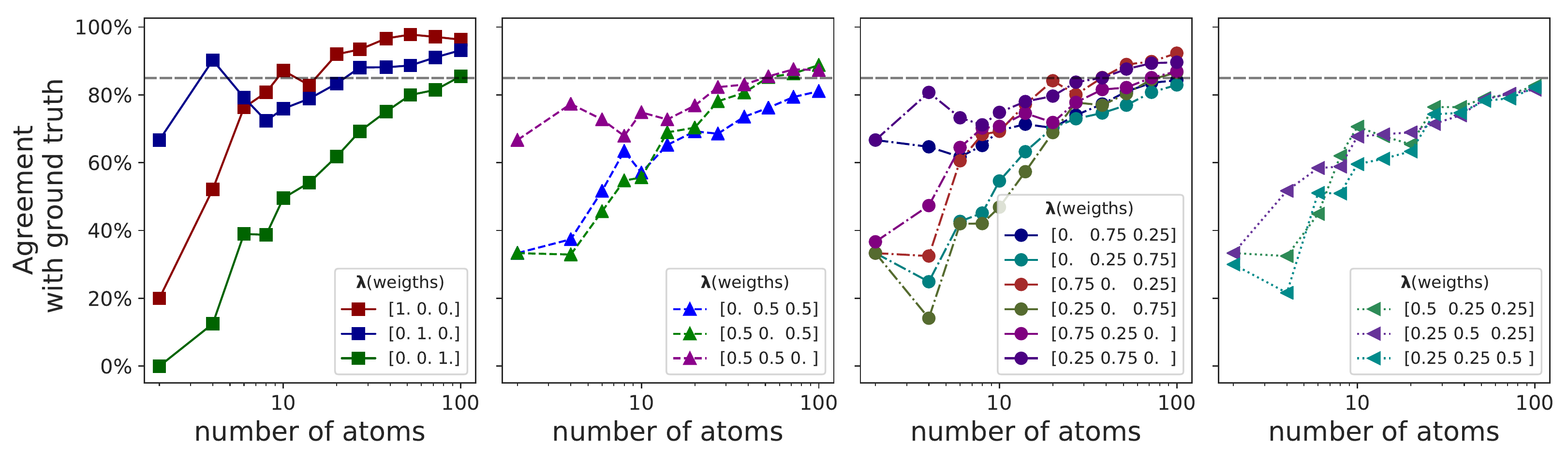}}\\
	\vspace{-3mm} 
	\subfloat[L-BW]{\includegraphics[width=0.95\linewidth]{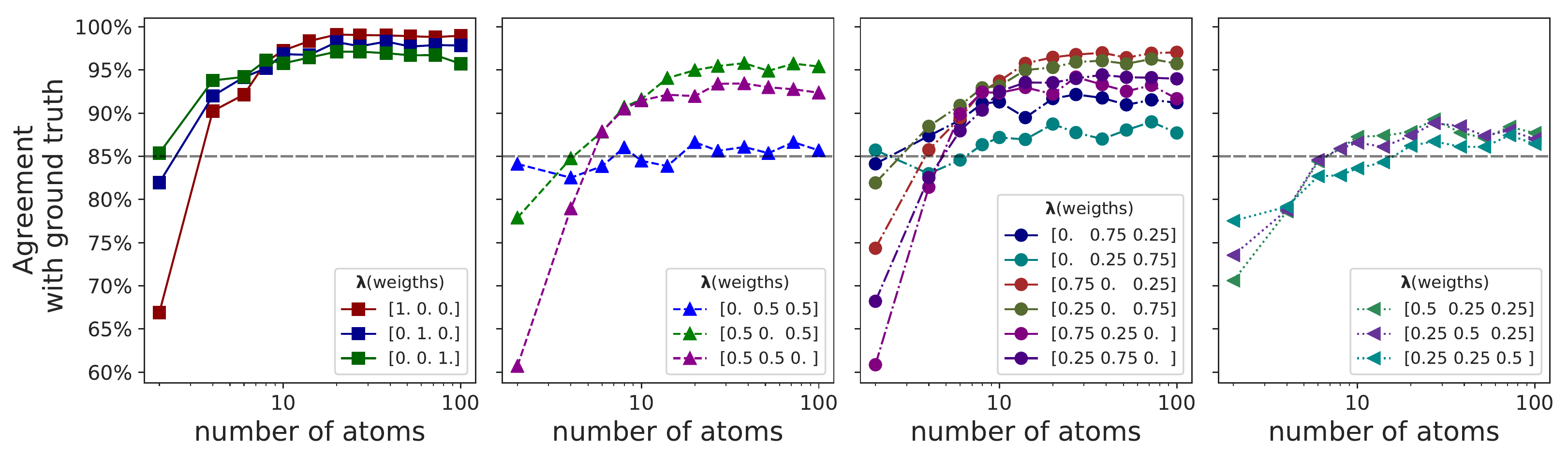}}
	\caption{\textbf{Support recovery on simulated data:}
		Evaluation of the performance varying the number $k$ of atoms, 
		for barycenters obtained with different weights $\blambda \in \Delta^{3}$.
		Coordinates in the Wasserstein simplex represent shapes: 
		cat ($\blambda = [1, 0, 0]$), 
		rabbit  ($\blambda = [0, 1, 0]$), and
		tooth ($\blambda = [0, 0, 1]$).
		The horizontal dashed line denotes $85\%$ of the agreement.
		%
	}
	\label{fig:2D_barycenter_quality}
\end{figure}

\subsection{Statistical-parity fairness experiment\label{sec:supp_1}}
\begin{figure}[H]
	\centering
		\includegraphics[width=0.3\linewidth]{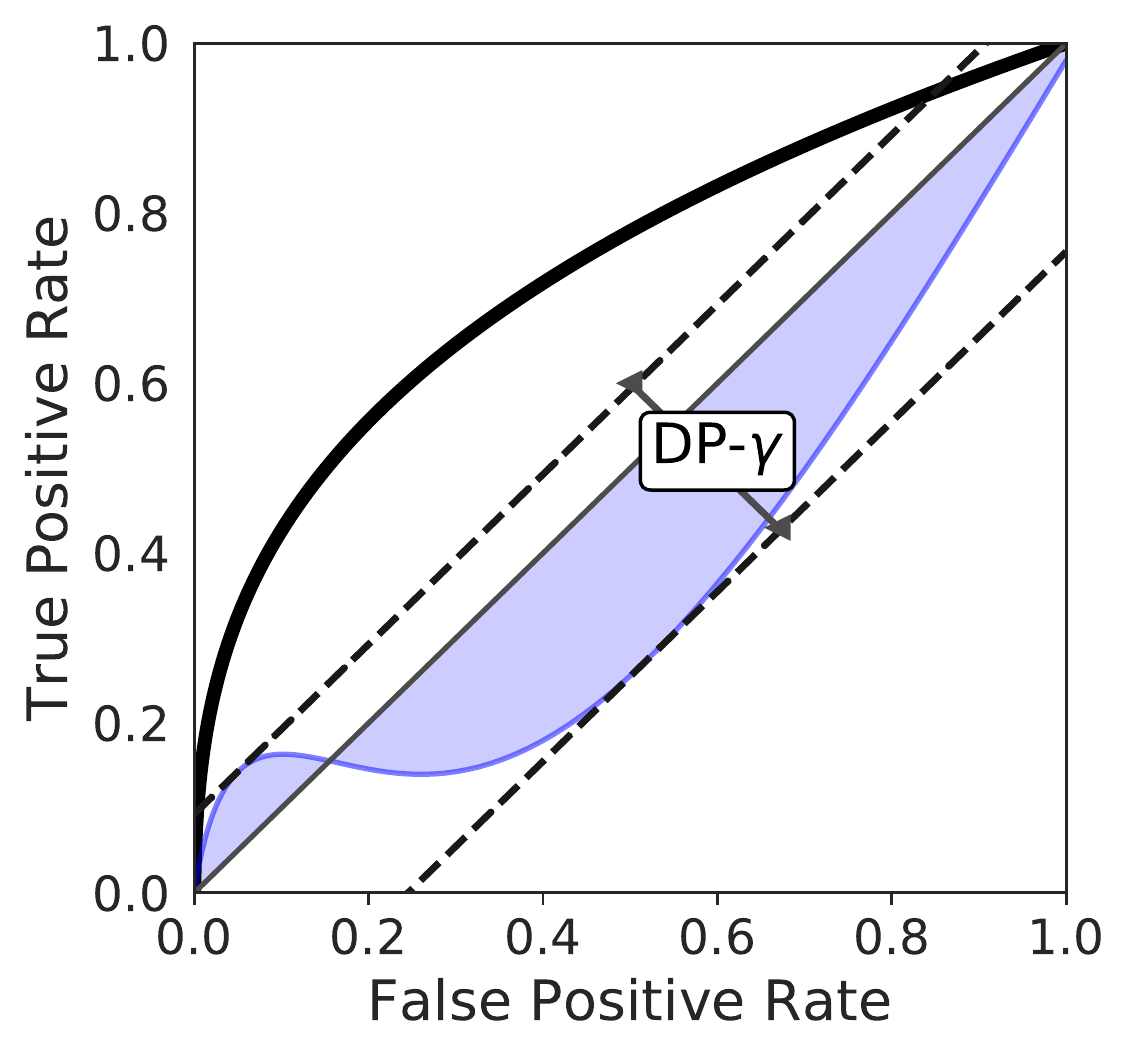}
	\caption{\textbf{Visual representation of the notion of fairness:} 
		The solid black line is the ROC curve for the $\mathbf{y}$ target.
		The solid blue line is the ROC curve for the protected $\mathbf{z}$ label that encodes the group membership of each sample $\bsx$.
		DP-$\gamma$ is the maximum distance of the latter from the diagonal, 
		which represents lack of predictability.}
	\label{fig:supp_dp_delta}
\end{figure}

\begin{figure}[t]
	\centering
	\includegraphics[width=0.75\linewidth, trim={0 10mm 0 0}, clip]{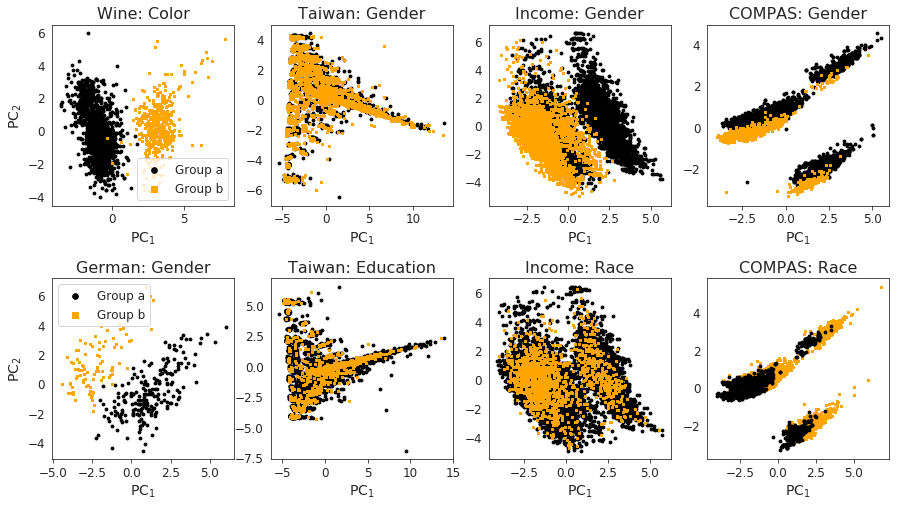}
	\caption{\textbf{Visualization of the first two principal components for different datasets:}
	We can see that for most of the datasets the groups are well separated. 
	Training a classifier on top of this representation will lead to leak of information about the protected groups.
 }
	\label{fig:supp_principal_components}
\end{figure}
\begin{figure}[t]
	\centering
	\small
	\includegraphics[width=0.75\linewidth,  trim={0 10mm 0 0}, clip]{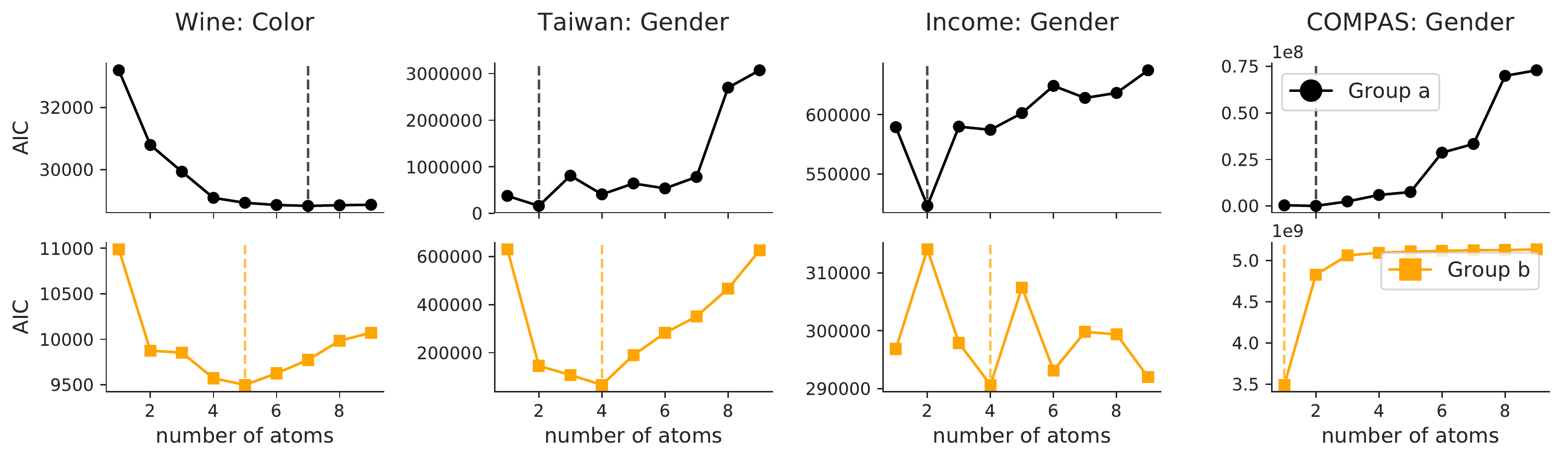}\\
	\includegraphics[width=0.75\linewidth,  trim={0 10mm 0 0}, clip]{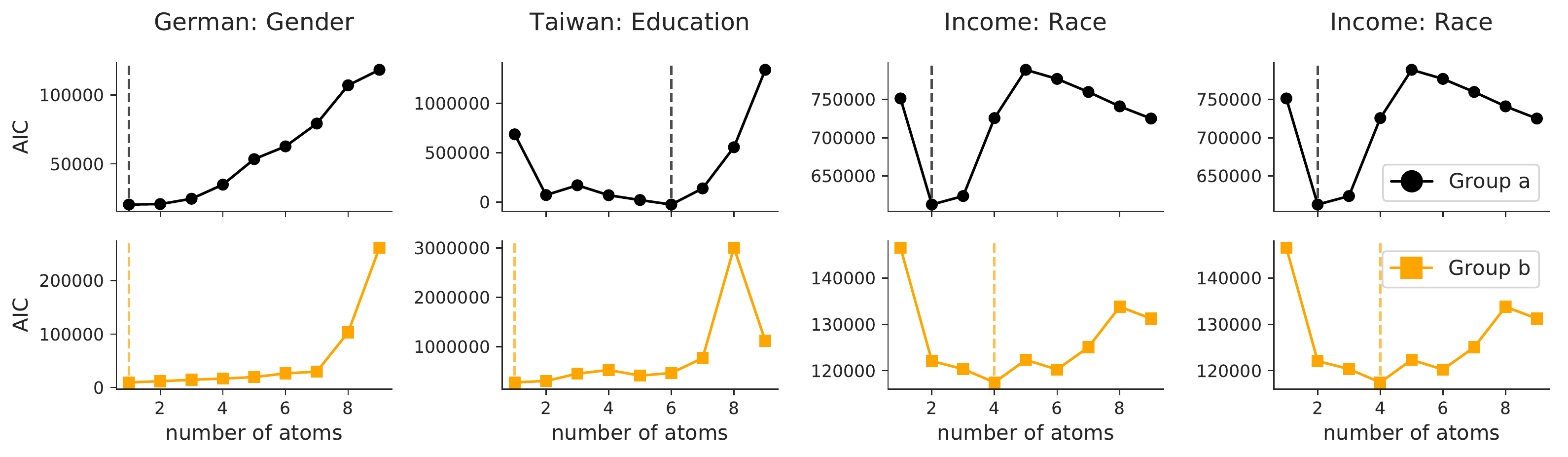}
	\caption{\textbf{Setting the number of components:}  The AIC as a function the number of Gaussian components for all datasets and sensitive conditions.
	For most of the datasets, the optimal number of components is different for each sensitive group.
}
	\label{fig:supp_gmm_components_datasets}
\end{figure}

\begin{figure}[t]
	\small
	\begin{center}
	\includegraphics[width=0.65\textwidth, trim={0 3mm 0 0}, clip]{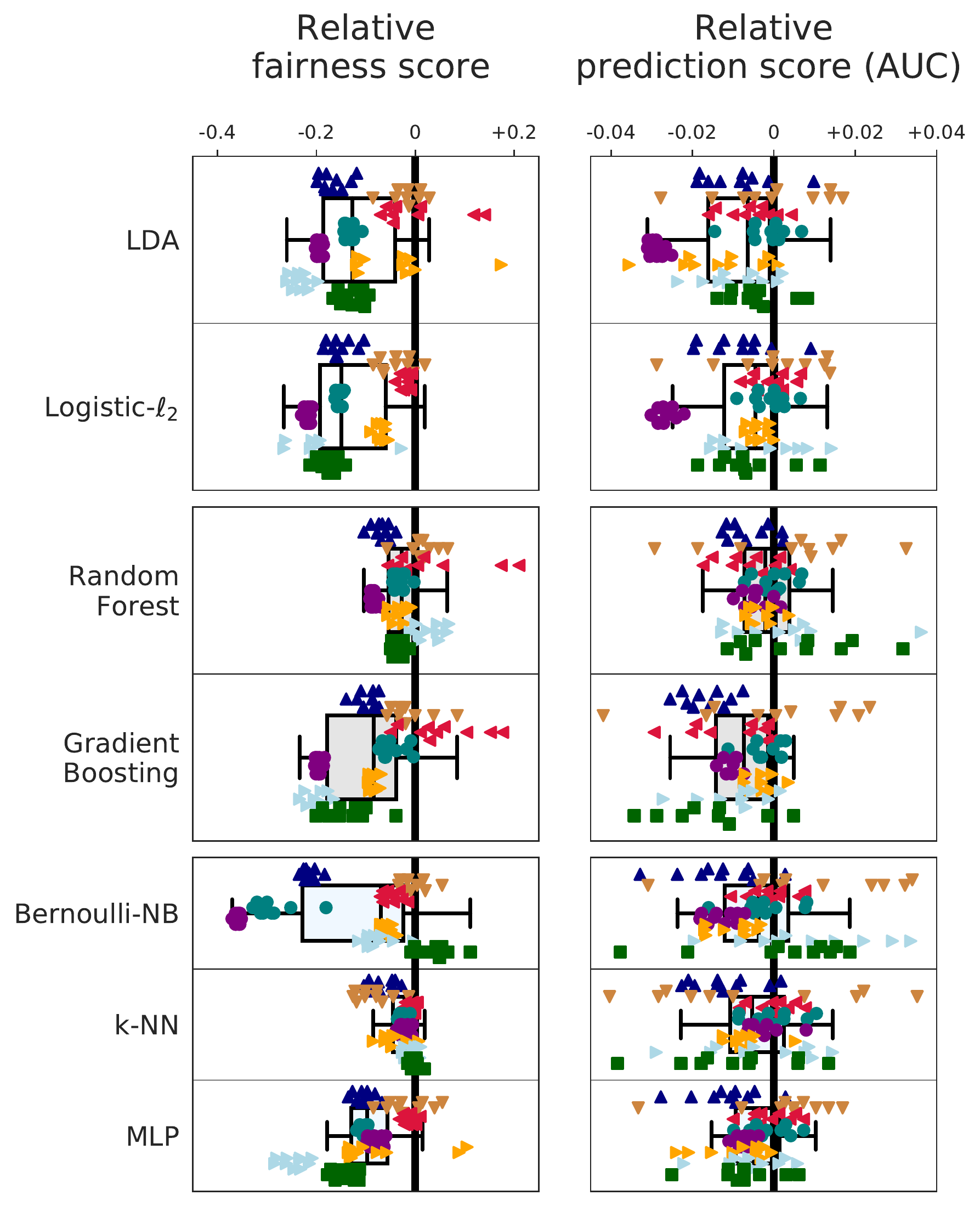}\\
	\includegraphics[width=0.5\linewidth]{legend_classification.pdf}
	\vspace{-3mm}
	\caption{\textbf{Impact of mixing distributions with L-BW on performance of various conventional classifiers:} 
		The performance of classifiers after mixing groups relative to 
		mean scores of the raw classifier. 
		The fairness score, the lower, the better ($< 0$).
		The AUC, the higher, the better ($> 0$).
		Mixing groups improves the fairness for most methods while preserving the predictive accuracy.
	}
	\label{fig:supp_relative_performance_plugin}
	\end{center}
\end{figure}
Fig.\ref{fig:supp_relative_performance_plugin} shows the impact of mixing groups by
 L-BW in various conventional classifiers: 
Logistic regression ($\ell_2$ penalty), 
Naive Bayes (Bernoulli), 
k-Nearest Neighbors (k-NN),
Linear Discriminant Analysis (LDA), 
Gradient Boosting Trees ($100$ trees), 
Random Forest ($100$ trees), and
Multi Layer Perceptron (MLP, $100$ hidden units and ReLu as activation function).
The fairness score (DP-$\gamma$) of K-NN and Random-Forest does not display any improvement, 
whereas the other classifiers present significant results ($\text{p}_\text{value} < 10^{-8}$, 
paired Wilcoxon rank test).
L-BW reduces the predictive performance of Gradient Boosting Trees and Linear Discriminant Analysis.
Contrary to other classifiers, where the difference to the raw classifier is not significant.

\end{document}